# A Generic Deep Architecture for Single Image Reflection Removal and Image Smoothing

Qingnan Fan*[1] Jiaolong Yang[2] Gang Hua[2] Baoquan Chen[1,3] David Wipf[2]
[1]Shandong University [2]Microsoft Research [3]Shenzhen Research Institute, Shandong University
fqnchina@gmail.com, {jiaoyan,davidwip,ganghua}@microsoft.com, baoquan@sdu.edu.cn

## Abstract

*This paper proposes a deep neural network structure that exploits edge information in addressing representative low-level vision tasks such as layer separation and image filtering. Unlike most other deep learning strategies applied in this context, our approach tackles these challenging problems by estimating edges and reconstructing images using only cascaded convolutional layers arranged such that no handcrafted or application-specific image-processing components are required. We apply the resulting transferrable pipeline to two different problem domains that are both sensitive to edges, namely, single image reflection removal and image smoothing. For the former, using a mild reflection smoothness assumption and a novel synthetic data generation method that acts as a type of weak supervision, our network is able to solve much more difficult reflection cases that cannot be handled by previous methods. For the latter, we also exceed the state-of-the-art quantitative and qualitative results by wide margins. In all cases, the proposed framework is simple, fast, and easy to transfer across disparate domains.*

## 1. Introduction

Inspired by the tremendous success of deep learning for large-scale visual recognition tasks like ILSVRC [28, 20], a variety of recent work has investigated deep neural networks for low-level computer vision tasks such as image denoising [27, 11], shadow removal [15], and image smoothing [38, 24]. Given that edges represent an important cue in addressing many of these problems, networks that can replace computationally-expensive or otherwise inflexible edge-aware filters naturally show promise.

For example, the underlying goal of image smoothing is to extract sparse salient structures, like perceptually important edges and contours, while minimizing the color differences in image regions with low amplitude. To approximate different edge-sensitive image smoothing filters which potentially have slow runtimes [2, 6, 8, 26, 37, 39, 42, 43] with deep networks, it has been proposed to first learn a salient gradient/weight map and then subsequently filter images via simpler, weighted optimization procedures [38] or iterative recursive processing techniques [24]. The above approaches focus on solving a single/major problem using a plain CNN model followed by more traditional, inflexible operations inspired by fixed filtering methods. Consequently, they are not fully extensible to implementing broader image smoothing effects or other significantly different problems such as image layer separation.

In this latter regard, one typical case where gradient domain statistics are relevant is in dealing with image reflections, that are often at least partially out of focus, when provided with a single image. When taking a photo through a glass window, the glare or reflection tends to distract the eye from the scene behind the glass. Many attempts to mitigate these effects, such as using a polarizer [19, 31], draping a large piece of black cloth over the lens and the glass to block ambient light from behind, or changing positions [22, 40, 41], are simply infeasible in many practical situations. Moreover, when taking photographs in airplane, museum, aquarium, or related environments, there is no other recourse but to shoot through the window. Consequently, it is common for photographers to simply widen the aperture of the camera and blur out the reflections.

To address this reflection removal problem from a computational perspective, traditional imaging models assume that the captured image $\mathbf{I}$ is a linear combination of a background layer $\mathbf{B}$ and a reflection layer $\mathbf{R}$, *i.e.*, $\mathbf{I} = \mathbf{B} + \mathbf{R}$. Obviously this is an ill-posed problem as there exist infinite feasible solutions, and hence most reflection removal algorithms require multiple input images [7, 30, 1, 19, 12, 22, 40, 41] or manual user interactions [21] to label reflection- and background-layer gradients, thus condensing the space of candidate solutions. However, one exploitable property in the reflection removal problem is that the gradients or perceptual structures of the two layers exhibit different distributions, since reflections often display a greater degree

*This work was done when Qingnan Fan was an intern at MSR.

of blurring. This then naturally leads us towards edge-based solutions, with data-driven network variants considered herein.

In this paper, we present a *Cascaded Edge and Image Learning Network (CEILNet)* that can be tailored to solve different image processing tasks such as layer separation (*e.g.*, reflection removal) and image filtering (*e.g.*, image smoothing). We rely on an overriding generic structure that is specialized in each instance via domain-specific edge information. The core framework operates in a very intuitive way. In brief, we separate the difficult task of directly predicting an image into two subproblems: (*i*) predicting the edge maps of the target images via a deeply supervised sub-network, and then (*ii*) reconstructing the target images by leveraging the predicted edge maps. These tasks are learned end-to-end by cascading two similar simple CNNs, and no hand-crafted modules are required. The edge map represents any color difference between each pair of adjacent pixels for task-specific target images, instead of sparse salient structures as in edge detection problems.

Of course, these objectives require ample training data to be feasible in practice. For image smoothing, this is not especially problematic provided sufficient computational resources are available for producing filter outputs across a corpus of images. However, for many layer separation tasks ground-truth instances are scarce. We therefore propose a novel weakly supervised learning method for training our reflection removal pipeline. This involves the use of images synthetically corrupted via reflections that mimic the physical properties of those found in natural scenes.

Our contributions can be summarized as follows:

- We propose a new, generic Cascaded Edge and Image Learning Network (CEILNet) that relies only on convolutional layers and is specifically designed to tackle edge-sensitive image processing tasks without resorting to any handcrafted, application-specific components. This structure is fast, extensible, and easy to reproduce, facilitating the seamless transfer to different low-level vision problems.
- We are the first to solve the challenging layer-separation problem of reflection removal from single images using deep learning techniques. We also propose a novel weakly supervised learning strategy combined with CEILNet.
- Beyond reflection removal, we demonstrated state-of-the-art visual and numerical performance using CEILNet on the image smoothing task, surpassing previous methods by a wide margin.

## 2. Related Work

**Reflection Removal:** Reflection removal is fundamentally an underdetermined problem and therefore requires prior knowledge or additional information to achieve any degree of success. Perhaps the most popular practical remedy is to use multiple input images, such as flash/non-flash image pairs [1], focus/defocus pairs [30], video sequences where background and reflection exhibit different motions [7, 34, 29, 9, 22, 33, 12, 40, 41], or those obtained through a polarizer at two or more orientations [19, 31, 29]. A few ambitious approaches attempt single image reflection removal, a far more difficult but practical scenario. In [21], manual annotation is required to guide an optimization-based layer separation. [32] compensates for the limited information by exploiting ghost cues, but this approach is not applicable beyond this somewhat specialized situation, or in the majority of practical cases. [35] leverages a multi-scale DoF computing strategy to separate reflection from background.

In terms of automatic reflection removal from a single image with minimal assumptions, the work most closely related to ours is [23]. This approach assumes the reflected layer is relatively blurry compared to the background scene, thus large gradients in it are strongly penalized in their optimization. However, we observe that the reflection in many real-world photographs, although indeed sometimes out of focus or blurry, is nonetheless produced by bright lights and often comprises the brightest portion of an image. The regional gradients associated with these reflections can therefore be quite large, violating the assumption in [23]. In this work, we synthesize a database of training samples that better capture the background and reflection statistics, and replace prior knowledge injected through explicit gradient penalization or energy minimization with a particular deep network to capitalize on this form of weak supervision. Empirically we will later show that indeed significant improvement is possible on real images.

**Image Smoothing:** Given the recent effectiveness of parallel computation through GPUs, and the strong learning capability of deep neural networks, replacing computationally-expensive, optimization-based smoothing filters with cheap neural modules has drawn a lot of attention [38, 24]. However, because accurately capturing smoothing effects with a fully convolutional deep network can be challenging, [38] trains a shallow CNN on the gradient domain followed by an optimized image reconstruction post-processing step with sensitive parameters tuned for each different smoothing filter. From a somewhat different perspective, by treating spatially-variant recursive networks as surrogates for a group of distinct filters, [24] combines sparse salient structure prediction implemented as CNN with image filtering in a hybrid neural network.

While significant differences exist, all of these prior methods lean on traditional optimization or filtering techniques at some point in their pipelines. Moreover, they are mostly applied to image smoothing using filter- or effect-dependent implementations without a universal, trainable parametric structure. This can potentially contribute to degraded performance since no single optimization or filtering strategy is likely to generalize to all different image smoothing effects. In contrast, our method learns a generic, fully-convolutional structure with no attendant postprocessing or otherwise fixed, filter-inspired structures. Empirical experiments demonstrate that this revised strategy outperforms the best existing work by a wide margin.

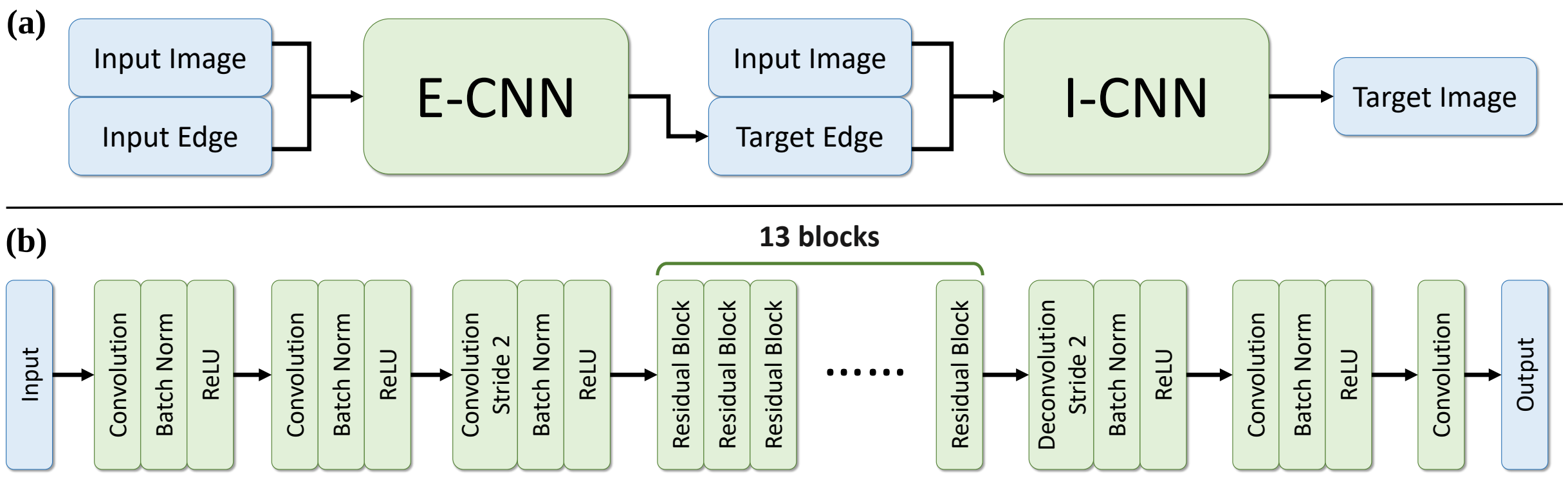

Figure 1. The proposed deep network architecture CEILNet. **(a)** The cascaded edge and image prediction pipeline. Two CNN networks, E-CNN and I-CNN, are used for edge prediction and image reconstruction, respectively. I-CNN takes the output of E-CNN as input, giving rise to an end-to-end and fully convolutional solution. **(b)** The detailed CNN structure shared by E-CNN and I-CNN.

## 3. Network Structure

Our network consists of two cascaded sub-networks: an edge prediction network E-CNN and an image reconstruction network I-CNN. Figure 1 is a schematic description of the architecture, which is unchanged for both the reflection removal and image smoothing applications.

### 3.1. E-CNN: The Edge Prediction Network

When dealing with edge-sensitive image processing tasks like reflection removal and image smoothing, edges-related cues are naturally leveraged by many existing algorithms [21, 22, 40, 10, 38]. Similarly, given a source image $\mathbf{I}^s$, we apply a CNN to learn an edge map $\mathbf{E}^t$ of the target image $\mathbf{I}^t$ (*i.e.*, the background layer for reflection removal or the smoothed image for image smoothing). Note that the goal is to predict the edges of the *target image*, not the *input image*, and it is crucial not to confuse this procedure with conventional edge detection [3, 36].

In this work, our edge map is not binary, as we empirically found binary edge maps are less informative for the subsequent image reconstruction. Instead, we designed a simple but effective edge representation: the mean absolute color difference between a center pixel and its four-connected neighbors. Specifically, the edge map $\mathbf{E}$ of an image $\mathbf{I}$ is computed by:

$$\mathbf{E}_{x,y} = \frac{1}{4}\sum_c \big(|\mathbf{I}_{x,y,c} - \mathbf{I}_{x-1,y,c}| + |\mathbf{I}_{x,y,c} - \mathbf{I}_{x+1,y,c}| + |\mathbf{I}_{x,y,c} - \mathbf{I}_{x,y-1,c}| + |\mathbf{I}_{x,y,c} - \mathbf{I}_{x,y+1,c}|\big) \quad (1)$$

where $x, y$ are the pixel coordinates and $c$ refers to the channels in the RGB color space.

In order to ease the computation, we augment the source image $\mathbf{I}^s$ with its edge map $\mathbf{E}^s$ as an additional channel for input. The intuition behind is simple: either a reflection-free background layer or an image smoothed via a filtering process can be viewed as "simplified" versions of the original source images, and their edge maps are roughly "attenuated" versions of the source image edge maps. We observed that such an augmentation can not only lead to better results but also significantly accelerate the convergence during training. In summary, E-CNN approximates the following function $f$:

$$\mathbf{E}^t = f(\mathbf{I}^s, \mathbf{E}^s) \quad (2)$$

### 3.2. I-CNN: The Image Reconstruction Network

The second sub-network, I-CNN, is designed to reconstruct the target image $\mathbf{I}^t$ by learning how to process the input image $\mathbf{I}^s$ given the target edge map $\mathbf{E}^t$ predicted by E-CNN. In other words, it approximates the following function $g$:

$$\mathbf{I}^t = g(\mathbf{I}^s, \mathbf{E}^t) \quad (3)$$

The input image and the target edge are combined to be a 4-channel tensor as input, similar to E-CNN, hence their shared use of the same overall structure. Additionally, in the context of the edge-based image reconstruction step of image smoothing tasks, the I-CNN serves as a multi-purpose, data-driven substitution for traditional fixed filtering operations or optimization-based postprocessing structures.

1. Train E-CNN and I-CNN in parallel, with loss functions of Eq. 4 and Eq. 5 respectively.
2. Jointly train (fine-tune) E-CNN and I-CNN end-to-end, with loss in Eq. 6.

Figure 2. Our two-phase network training algorithm.

### 3.3. Details of CNN Layers

For simplicity, we employ the deep CNN structure shown in Fig. 1 (b) for both E-CNN and I-CNN. The two sub-nets only differ in the channel number of the final output, *i.e.*, 1 for E-CNN *vs.* 3 for I-CNN. In each case, we employ 32 convolutional layers with the same 3×3 kernel size (except for the third-to-last layer; see below). The intermediate 30 convolutional layers all have 64-dimensional input and output feature maps. The first 31 layers are followed by batch normalization (BN) and ReLU. To ensure better contextual information, we enlarge the receptive field by downsampling the internal feature map to half size and then upsampling it back by changing the stride of the third convolution layer to 2 and third-to-last convolution layer to deconvolution with stride 2 and kernel size 4×4. In this way, the receptive field is effectively enlarged without losing too much image detail, and meanwhile the computation cost is halved. For better performance and faster convergence, we implement the middle 26 convolution layers as 13 residual units [14] similar to [5].

Finally, to resolve the color attenuation issue [16, 17] observed in deep networks, we slightly magnify the predicted image $\mathbf{I}^t$ via $s_c \triangleq \arg\min_{s_c} ||\mathbf{I}_c^s - s_c \cdot \mathbf{I}_c^t||_2^2$ and $\mathbf{I}_c^t \leftarrow s_c \cdot \mathbf{I}_c^t$. This global color correction is implemented as a parameter-free layer after I-CNN. Its computational cost is negligible.

## 4. Network Training

This section first presents our training pipeline, that applies independently of the data source. Later we describe application-specific means of generating training samples.

### 4.1. Training Details

We employ a two-phase network training algorithm shown in Fig. 2. Specifically, we first train the sub-networks separately with ground-truth images and their edge maps to ensure the best individual performances. We then fine-tune the entire network end-to-end, granting the two sub-nets more opportunity to cooperate accordingly.

The sub-nets are trained by minimizing the mean squared errors (MSE) of their predictions. Let the symbol $*$ denote ground truth, the loss for edge prediction is

$$l_E(\theta) = ||\mathbf{E}^t - \mathbf{E}^{t*}||_2^2. \tag{4}$$

For image prediction, we minimize not only the color MSE but also the discrepancy of gradients:

$$\begin{aligned} l_I(\theta) &= \alpha\, ||\mathbf{I}^t - \mathbf{I}^{t*}||_2^2 \\ &+ \beta\, (||\nabla_x \mathbf{I}^t - \nabla_x \mathbf{I}^{t*}||_1 + ||\nabla_y \mathbf{I}^t - \nabla_y \mathbf{I}^{t*}||_1). \end{aligned} \tag{5}$$

The gradient discrepancy cost, though seemingly redundant, helps to prevent the deep convolutional network from generating blurry images [25]. In the joint training phase, we train the entire network by minimizing the loss:

$$l(\theta) = l_I(\theta) + \gamma\, l_E(\theta). \tag{6}$$

For all experiments across reflection removal and image smoothing, the loss coefficients are empirically set as $\alpha = 0.2$, $\beta = \gamma = 0.4$ (other selections produce similar results).

We initialize the convolution weights using the approach from [13] and train all networks using ADAM [18] with mini-batch size fixed at 1. When training the two sub-nets separately, the learning rate is set to 0.01 over the initial iterations, *e.g.*, 40 and 25 epochs for reflection and imaging smoothing tasks respectively. The entire network is then fine-tuned with the learning rate reduced to 0.001.

### 4.2. Training Data Generation

**Reflection Image Synthesis:** Real images with ground truth background layers are difficult to obtain. To generate enough training data, simply mixing two images with different coefficients (such as 0.8 for background and 0.2 for reflection) seems to be a straightforward and plausible compromise. Indeed, this strategy has been widely used in previous works [34, 29, 12, 23, 41] for analysis and quantitative evaluation. However, we found that networks trained on such images generalize poorly to real photographs. We therefore propose a novel synthesis method to better approximate real-world reflection.

As previously mentioned, we assume that the reflection is somewhat blurry relative to the background layer, which tends to be more sharp and clear. This is a valid assumption for many cases, as the camera is usually focused on the background target. Moreover, a photographer can easily widen the camera's aperture and blur out the reflections. A similar assumption is used by [23].

We expand on this assumption using a simple complementary observation. First, according to the Fresnel equation, we know that when incident light travels across media with different refractive indices (*e.g.*, glass and air) in front of some scene of interest, a portion of that light will be reflected back to the image plane. However, the actual visibility of this reflected light to the human eye or a camera depends on the relative intensity of light transmitted from the background scene. Therefore we may expect that only portions of the background layer transmitting modest light will be appreciably obstructed via a reflection layer, even if the latter is uniformly present across a scene. And yet in regions where reflections are apparent, their intensity can still

Randomly pick two natural images normalized to $[0,1]$ as background $\mathbf{B}$ and reflection $\mathbf{R}$ respectively, then:

1. $\tilde{\mathbf{R}} \leftarrow gauss_blur_{\sigma}(\mathbf{R})$ with $\sigma \sim \mathcal{U}(2,5)$
2. $\mathbf{I} \leftarrow \mathbf{B} + \tilde{\mathbf{R}}$
3. $m \leftarrow mean(\{\mathbf{I}(\mathbf{x},c) \mid \mathbf{I}(\mathbf{x},c) > 1, \forall \mathbf{x}, \forall c = 1,2,3\})$
4. $\tilde{\mathbf{R}}(\mathbf{x},c) \leftarrow \tilde{\mathbf{R}}(\mathbf{x},c) - \gamma \cdot (m-1), \forall \mathbf{x}, \forall c$; $\gamma$ set as $1.3$
5. $\tilde{\mathbf{R}} \leftarrow clip_{[0,1]}(\tilde{\mathbf{R}})$
6. $\mathbf{I} \leftarrow clip_{[0,1]}(\mathbf{B} + \tilde{\mathbf{R}})$

Output $\mathbf{I}$ as the synthesized image with $\mathbf{B}$ as the ground-truth background layer.

Figure 3. Reflection image data synthesis for weakly-supervised learning. The subtraction and clipping operators allow for reflection intensities that can saturate and vanish in various regions.

be arbitrarily large (even if partially blurred) and so a purely additive model with a weakly scaled reflection component is not always physically plausible.

Based on the above observations, we develop a new method summarized in Fig. 3 to synthesize images with realistic background and reflection layers. One key difference from naive image mixing is that the brightness overflow issue is avoided not by scaling down the brightness, but by subtracting an adaptively computed value followed by clipping. In this way: (*i*) reflection-free regions are very likely to appear which is consistent with natural images, (*ii*) strong reflections can occur in other places, and (*iii*) the reflection contrast is better maintained. Also note that we randomly pick the $\sigma$ of the Gaussian blur kernel between $[2,5]$, in contrast to a fixed large value ($\sigma = 5$) tested in [23]. We are interested in handling a wider range of real cases, including cases with lesser blurry reflections. Figure 6 (top) displays 4 synthetic images generated by our method, and Fig. 5 shows a result comparison with naive image mixing. For more comparisons and details regarding the synthesis process, see the *supplemental material*.

Note that synthetically generated samples serve as a form of weak supervision, as we ultimately deploy the trained model on new real images containing natural reflections.

**Generation of Smoothed Images:** For image smoothing, our network is trained to approximate the effect of existing filters. The training and testing data will simply be the smoothed images generated by applying those filters to existing image databases. Various filters are tested in Sec. 5.

# 5. Experiments

This section first presents self-comparison experiments to analyze the importance of proposed network architecture design choices. We then evaluate the full CEILNet against the state-of-the-art algorithms on the single-image reflection removal and image smoothing tasks.

Table 1. Result comparison for the image smoothing task (learning an $L_0$ filter [37]). CEILNet outperformed Domain Transform (DT) [10] and simple I-CNNs without E-CNN by large margins.

| | MSE | PSNR | SSIM |
|---|---|---|---|
| DT + input image edge | 124.41 | 27.38 | 0.806 |
| DT + pred. edge by E-CNN | 51.26 | 31.17 | 0.964 |
| DT + GT edge | 45.67 | 31.66 | 0.971 |
| I-CNN only | 37.79 | 32.58 | 0.969 |
| I-CNN only (64 layers) | 31.86 | 33.33 | 0.973 |
| I-CNN with input edge (64 layers) | 22.50 | 34.86 | 0.979 |
| CEILNet | **13.34** | **37.10** | **0.989** |

## 5.1. Network Analysis

For simplicity, our analysis will be mainly based on the representative results of approximating $L_0$ smoothing [37]. These results were obtained on 100 PASCAL VOC test images (refer to Sec. 5.3 for training and testing details).

**Is the target edge map from E-CNN helpful?** To verify the importance of the target edge map for image reconstruction, we removed E-CNN and trained a simple I-CNN model without the predicted target edge or replacing the predicted target edge with the input image edge. Table 1 shows that I-CNN with predicted edge (*i.e*., our CEILNet) outperformed I-CNN alone and I-CNN with input edge by significant margins, demonstrating the importance of target edge prediction. A visual comparison is shown in Fig. 4.

Similar results were obtained for reflection removal: the predicted background edges were found to be helpful for layer separation. Figure 5 shows a typical example.

**Does simply stacking more layers in I-CNN suffice?** Ideally, with enough depth, one may expect the network to handle target edge prediction implicitly without the need for an explicit E-CNN. We tried training a simple I-CNN with more convolutional layers. and found that the performance gets saturated quickly after more than 50 layers (a detailed figure is deferred to the supplementary material). Our CEILNet, *i.e*., 32-layer E-CNN + 32-layer I-CNN, achieved much better results than a 64-layer simple I-CNN (as shown in Table 1) and a best-performing 70-layer one (PSNR 33.37 *vs*. 37.10 by CEILNet).

**Is I-CNN better than a traditional method?** To answer this question, we replaced I-CNN with the Domain Transform (DT) technique [10]. The predicted target edge map by E-CNN and the input image are fed to DT to output smooth images. We also tried the ground-truth target edge and the input image edges. Table 1 shows that I-CNN with predicted edge from E-CNN (*i.e*., our CEILNet) outperformed all DT results by large margins. A visual comparison is presented in Fig. 4.

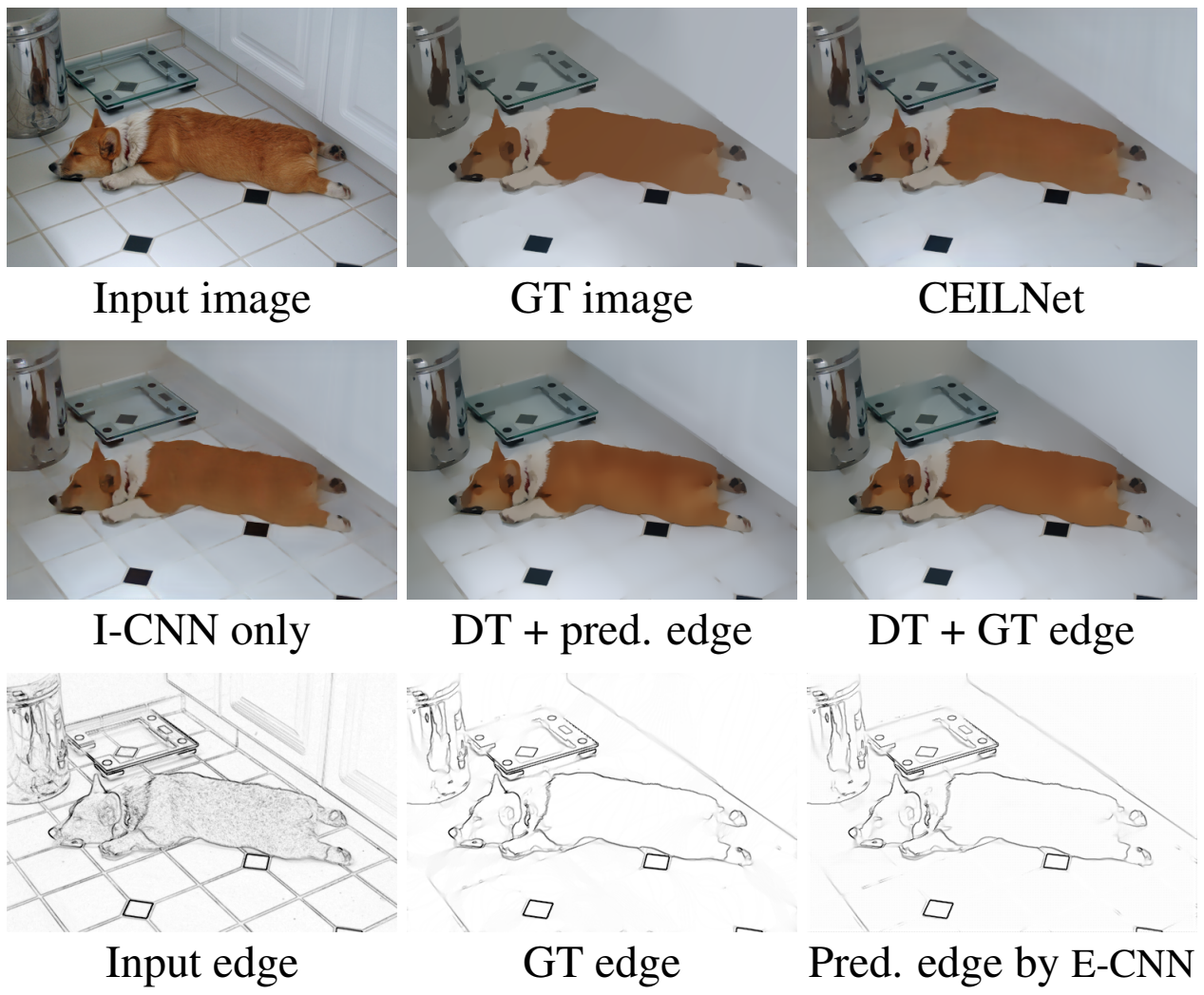

Figure 4. Qualitative comparison for the image smoothing task (learning an $L_0$ filter [37]). Our CEILNet generates a more satisfactory result than a simple I-CNN without E-CNN and than Domain Transform [10]. **Best viewed on screen with zoom.**

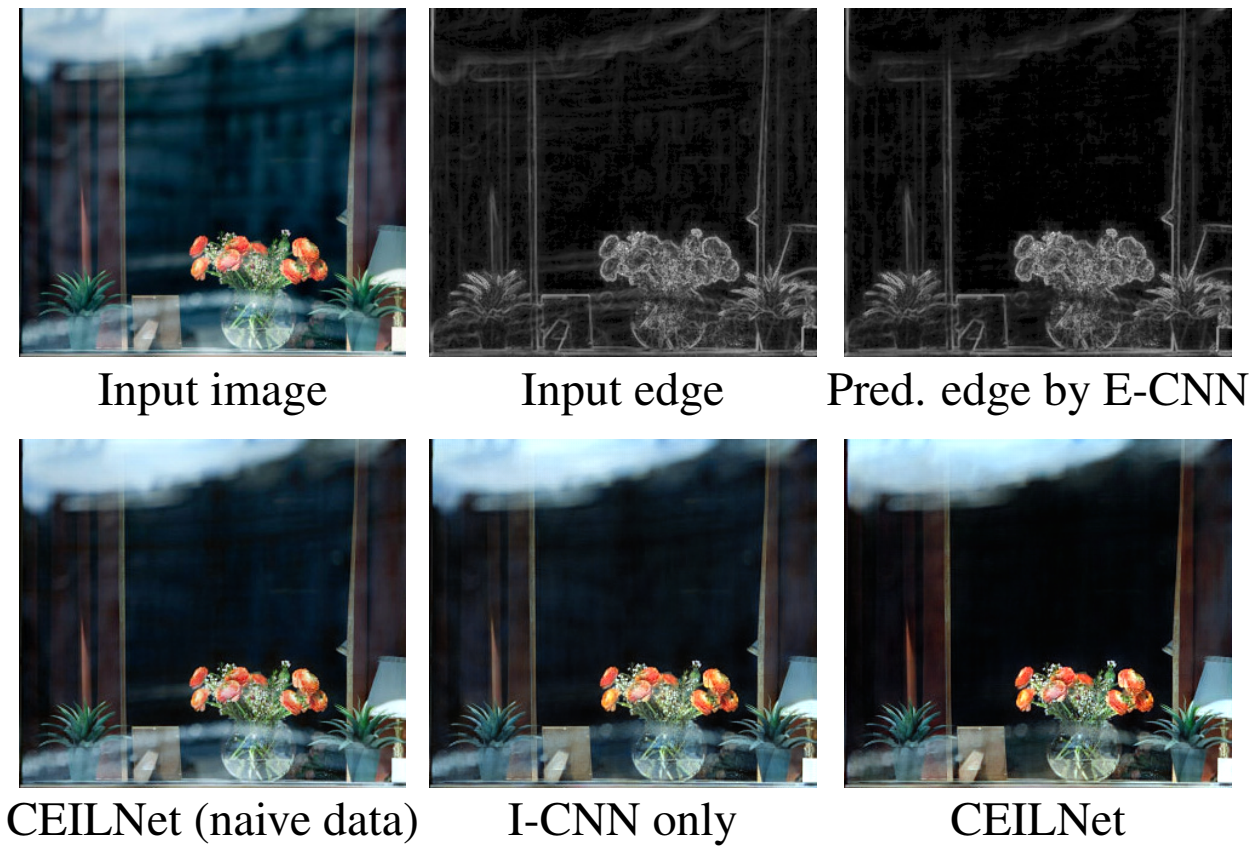

Figure 5. Qualitative reflection removal results on a real image. Our CEILNet removes more reflection and generates a clearer background image than a simple I-CNN without E-CNN, and than CEILNet trained with a naive image mixing strategy for data generation. **Best viewed on screen with zoom.**

For reflection removal, we also tried applying the layer separation algorithm in [22] with our predicted edges as input, but no satisfactory results were obtained.[1]

### 5.2. Reflection Removal

**Training Data:** We applied the method described in Sec. 4.2 to synthesize training data for the reflection removal task to accommodate our weakly-supervised learning pipeline. We used 17K natural images from the PASCAL VOC dataset [4] for the synthesis. These images were collected from Flickr, and represent a wide range of viewing conditions. Two natural images were used to generate one synthetic image containing a background layer and a reflection layer, resulting in 8.5K synthetic images in total. We split these images into a training set of 7,643 images and a test set with 850 images for quantitative comparison. The training images are also cropped to 224×224. The algorithm described in Fig. 2 was then applied, and we did not observe over-fitting in any of the training sub-tasks.

[1][22] utilizes multiple images to identify background edges, which are used as prior to guide layer septation. Their septation algorithm did not work well with our edge maps as it assumes non-blurry reflections and requires binary edge maps.

Table 2. Quantitative comparison of our method with Li and Brown [23] on 100 synthetic images with reflection.

| PSNR | | SSIM | |
|---|---|---|---|
| [23] | Ours | [23] | Ours |
| 15.50 | **18.55** | 0.786 | **0.857** |

**Method Comparison:** We tested our CEILNet against the state-of-the-art, single-image approach from [23]. For a quantitative comparison, we randomly selected 100 images in our test dataset, and evaluate the PSNR and SSIM metrics for the predicted $\mathbf{B}$ from both algorithms. The default parameters of [23] were used for evaluation. Table 2 shows that CEILNet significantly outperformed [23].

Figure 6 presents some qualitative results of our method compared against [23] on both synthetic and real images. The reflection image estimates are computed via $\mathbf{R} = \mathbf{I} - \mathbf{B}$. We tuned the parameters of [23] for each image to get the best visual result. It can be seen that [23] tends to generate a blurry reflection layer with brightness covering the whole image. It largely failed to remove less blurry, high contrast or partially present reflections. This is because [23] employs strong priors to penalize abrupt color transitions in $\mathbf{R}$ which, however, may be common in real cases. In contrast, our CEILNet is able to separate out the reflections reasonably well even if some of them are very bright and shiny, and without jeopardizing the reflection-free regions. More results and comparisons are deferred to the *supplementary material* due to space limitation.

### 5.3. Image Smoothing

**Training Data:** For image smoothing, we used the 17K natural images in the PASCAL VOC dataset as input, and generated the filtered images using existing image smoothing algorithms as the ground truth. These images are fed to the network without cropping. We also randomly pick 100 images in the PASCAL VOC dataset for testing. We again use the algorithm in Fig. 2 to train our CEILNet.

**Method Comparison:** We tested 8 image smoothing algorithms for the network to approximate, including bilateral filter (BLF) [26], iterative bilateral filter (IBLF) [8], rolling guidance filter (RGF) [42], RTV texture smoothing (RTV) [39], weighted least square smoothing (WLS) [6],

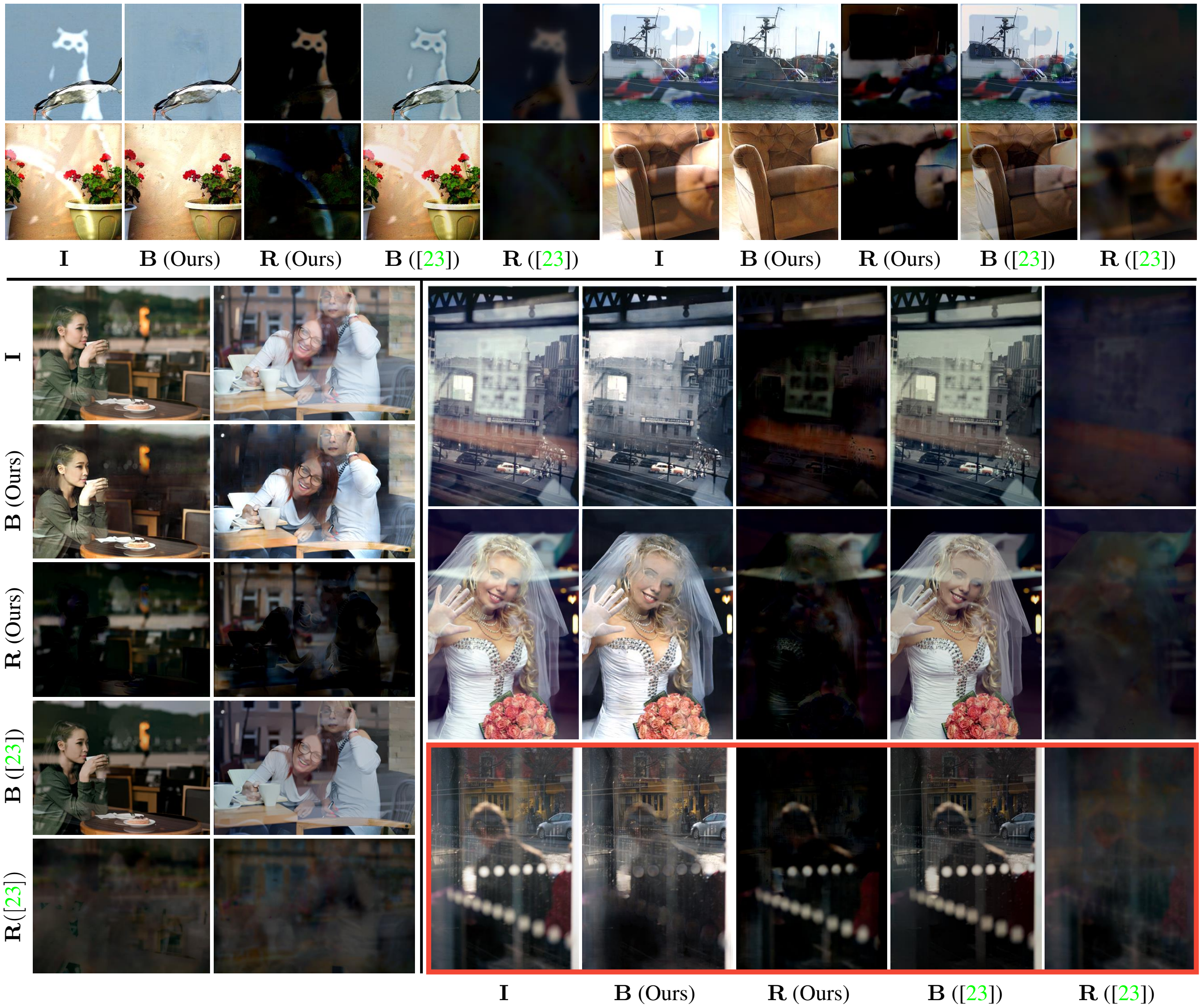

Figure 6. Qualitative results of the single image reflection removal task on synthetic (*top two rows*) and real (*bottom rows*) images. Visually inspected, our method can largely remove the reflection and produce reasonably good background images under various situations. The method of Li and Brown [23] clearly underperformed. The last example is a partial failure case for our method due to the strong reflection and weak transmitted light, but still the result is superior to [23]. **Best viewed on screen with zoom.**

Table 3. Quantitative comparison on the image smoothing tasks. We report the PSNR and SSIM metrics (larger is better) for 8 different smoothing filters, and compare our method with Xu *et al*. [38]. Average values are computed with the preceding 7 cases.

| | | BLF | IBLF | $L_0$ | RGF | RTV | WLS | WMF | $L_1$ | Ave. |
|---|---|---|---|---|---|---|---|---|---|---|
| PSNR | [38] | 35.02 | 32.97 | 31.66 | 32.49 | 35.68 | 33.92 | 29.62 | | 32.62 |
| | Ours | **43.76** | **38.18** | **37.10** | **42.05** | **44.03** | **41.39** | **39.70** | **36.99** | **40.40** |
| SSIM | [38] | 0.976 | 0.962 | 0.966 | 0.950 | 0.974 | 0.963 | 0.960 | | 0.964 |
| | Ours | **0.995** | **0.989** | **0.989** | **0.991** | **0.994** | **0.994** | **0.989** | **0.982** | **0.990** |

Table 4. Running time comparison (in seconds). We compare the running time of our method against different traditional methods as well as deep learning based methods of Xu *et al*. [38] and Liu *et al*. [24] at various resolutions.

| | BLF | IBLF | RGF | $L_0$ | WMF | RTV | WLS | $L_1$ | [38] | [24] | Ours |
|---|---|---|---|---|---|---|---|---|---|---|---|
| QVGA (320×240) | 0.03 | 0.11 | 0.22 | 0.17 | 0.62 | 0.41 | 0.70 | 32.18 | 0.23 | 0.07 | 0.008 |
| VGA (640×480) | 0.12 | 0.40 | 0.73 | 0.66 | 2.18 | 1.80 | 3.34 | 212.07 | 0.76 | 0.14 | 0.009 |
| 720p (1280×720) | 0.34 | 0.97 | 1.87 | 2.43 | 4.98 | 5.74 | 13.26 | 904.36 | 2.16 | 0.33 | 0.010 |

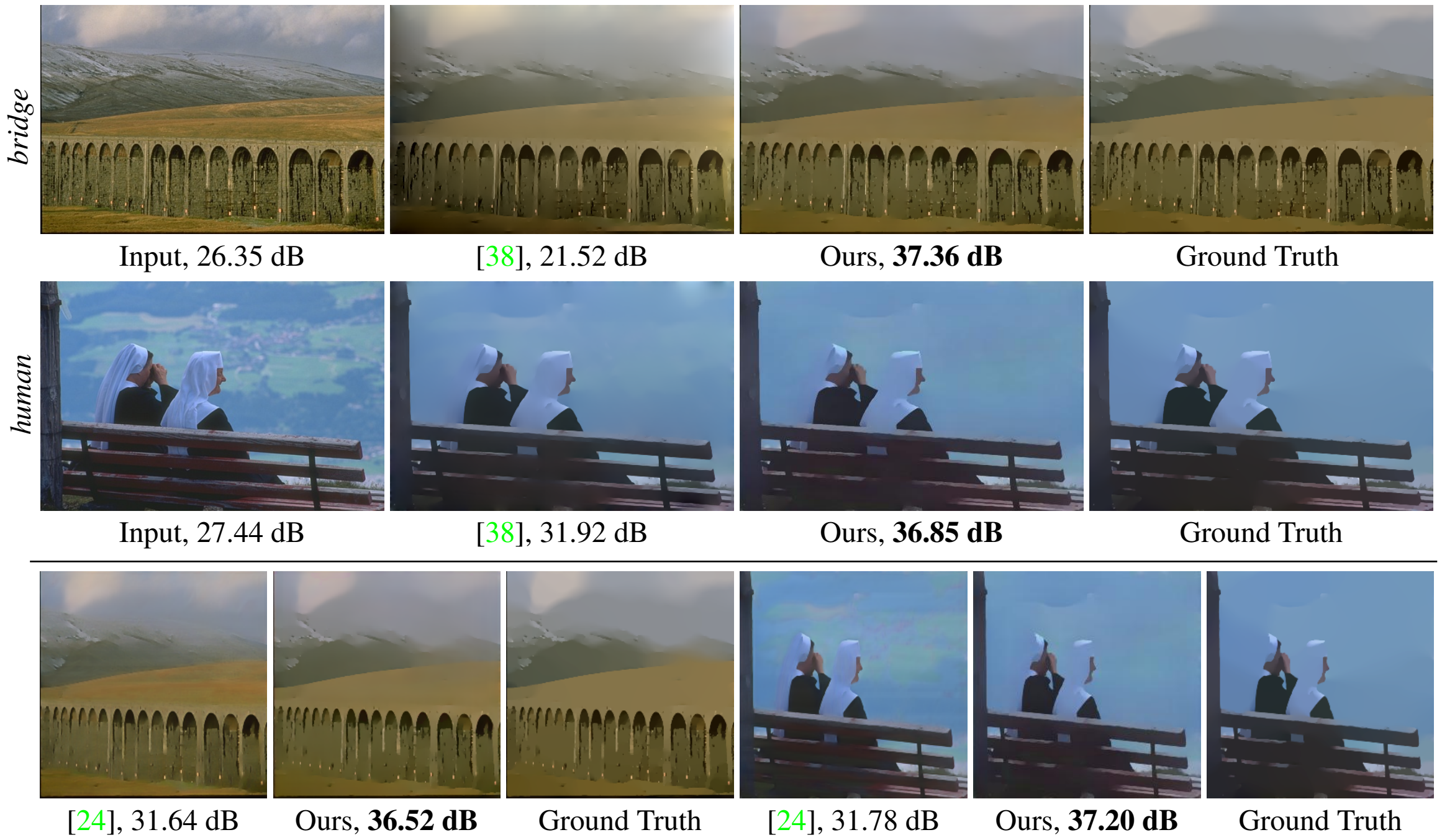

Figure 7. Qualitative results on the image smoothing task. All the methods are trained to approximate $L_0$ smoothing [37]. Top: Comparison with Xu *et al*. [38]. Bottom: Comparison with Liu *et al*. [24] on the 256×256 image size. Our results are visually much closer to the ground truth. The numbers show the PSNR values. **Best viewed on screen with zoom.**

Table 5. Comparison with Liu *et al*. [24] on image smoothing.

| | PSNR | | SSIM | |
|---|---|---|---|---|
| | [24] | Ours | [24] | Ours |
| $L_0$ | 32.26 | **36.62** | 0.958 | **0.986** |
| RGF | 38.64 | **40.80** | 0.986 | **0.989** |
| WLS | 38.29 | **40.27** | 0.983 | **0.992** |
| WMF | 33.29 | **37.75** | 0.951 | **0.986** |
| Ave. | 35.64 | **39.36** | 0.966 | **0.988** |

weighted median filter (WMF) [43], $L_0$ smoothing [37] and $L_1$ smoothing [2].

Table 3 presents the quantitative results of our method and [38] on the test set with 100 images. In can been seen that our network achieved much better results than [38] for all the 8 filters, on both the PSNR and SSIM metrics. We also compare our results with [24], whose models for 4 filters are publicly available. Note that at the time of writing, the latest code of [24] released by their authors cannot run on arbitrary image size due to some implementation constraints, so we use their default size of 256×256. Table 5 shows that our method also significantly outperformed [24] for all the 4 filtering algorithms.

Figure 7 presents two visual results of our method compared to others. It can be observed that the method of [24] generated obvious artifacts compared to the ground truth for both two cases, while [38] produced some unwanted color transitions in the right and bottom left regions of the "bridge" image, resulting in a PSNR even lower than the raw input image. In contrast, our results are visually more close to the ground truth. More results and discussions can be found in the *supplementary material*.

**Running Time:** We evaluate the running time of the eight traditional smoothing algorithms and the three deep learning based methods with respect to different image sizes on the same computer (NVIDA DGX-1). Table 4 shows that our method[2] runs faster than others in most of the cases. It can approximate any traditional algorithm at about 100 fps for 1280×720 images.

## 6. Conclusions and Future Work

We have proposed CEILNet, a generic deep architecture for edge-sensitive image processing. We provided the first learning-based solution to the challenging single image reflection removal problem using CEILNet and with the aid of a novel reflection image synthesis method. We have also significantly advanced the state-of-the-art in DNN-based image smoothing. Our future work includes testing CEILNet on more image processing tasks. Promising results for image denosing and inpainting have been obtained in our preliminary experiments.

**Acknowledgement** This work was partially supported by National 973 Program (2015CB352501), Shenzhen Innovation Program (JCYJ20150402105524053).

[2]The running time of our method reported in the previous ICCV version included the time for the data generation process. Here we exclude that part for a fair comparison.

## References

[1] A. Agrawal, R. Raskar, S. K. Nayar, and Y. Li. Removing photography artifacts using gradient projection and flash-exposure sampling. *ACM Transactions on Graphics (TOG)*, 24(3):828–835, 2005. 1, 2

[2] S. Bi, X. Han, and Y. Yu. An $L_1$ image transform for edge-preserving smoothing and scene-level intrinsic decomposition. *ACM Transactions on Graphics (TOG)*, 34(4):78, 2015. 1, 8

[3] J. Canny. A computational approach to edge detection. *IEEE Transactions on Pattern Analysis and Machine Intelligence (T-PAMI)*, (6):679–698, 1986. 3

[4] M. Everingham, L. Van Gool, C. K. Williams, J. Winn, and A. Zisserman. The pascal visual object classes (voc) challenge. *International Journal of Computer Vision (IJCV)*, 88(2):303–338, 2010. 6

[5] Q. Fan, D. Wipf, G. Hua, and B. Chen. Revisiting deep image smoothing and intrinsic image decomposition. *arXiv preprint. arXiv:1701.02965*, 2017. 4

[6] Z. Farbman, R. Fattal, D. Lischinski, and R. Szeliski. Edge-preserving decompositions for multi-scale tone and detail manipulation. *ACM Transactions on Graphics (TOG)*, 27(3), 2008. 1, 6

[7] H. Farid and E. H. Adelson. Separating reflections and lighting using independent components analysis. In *IEEE Conference on Computer Vision and Pattern Recognition (CVPR)*, volume 1, pages 262–267, 1999. 1, 2

[8] R. Fattal, M. Agrawala, and S. Rusinkiewicz. Multiscale shape and detail enhancement from multi-light image collections. *ACM Transactions on Graphics (TOG)*, 26(3), 2007. 1, 6

[9] K. Gai, Z. Shi, and C. Zhang. Blind separation of superimposed moving images using image statistics. *IEEE Transactions on Pattern Analysis and Machine Intelligence (T-PAMI)*, 34(1):19–32, 2012. 2

[10] E. S. Gastal and M. M. Oliveira. Domain transform for edge-aware image and video processing. In *ACM Transactions on Graphics (TOG)*, volume 30, page 69. ACM, 2011. 3, 5, 6

[11] M. Gharbi, G. Chaurasia, S. Paris, and F. Durand. Deep joint demosaicking and denoising. *ACM Transactions on Graphics (TOG)*, 35(6):191, 2016. 1

[12] X. Guo, X. Cao, and Y. Ma. Robust separation of reflection from multiple images. In *IEEE Conference on Computer Vision and Pattern Recognition (CVPR)*, pages 2187–2194, 2014. 1, 2, 4

[13] K. He, X. Zhang, S. Ren, and J. Sun. Delving deep into rectifiers: Surpassing human-level performance on imagenet classification. In *IEEE International Conference on Computer Vision (ICCV)*, pages 1026–1034, 2015. 4

[14] K. He, X. Zhang, S. Ren, and J. Sun. Deep residual learning for image recognition. In *IEEE Conference on Computer Vision and Pattern Recognition (CVPR)*, pages 770–778, 2016. 4

[15] S. H. Khan, M. Bennamoun, F. Sohel, and R. Togneri. Automatic shadow detection and removal from a single image. *IEEE Transactions on Pattern Analysis and Machine Intelligence (T-PAMI)*, 38(3):431–446, 2016. 1

[16] J. Kim, J. Kwon Lee, and K. Mu Lee. Accurate image super-resolution using very deep convolutional networks. In *IEEE Conference on Computer Vision and Pattern Recognition (CVPR)*, pages 1646–1654, 2016. 4

[17] J. Kim, J. Kwon Lee, and K. Mu Lee. Deeply-recursive convolutional network for image super-resolution. In *IEEE Conference on Computer Vision and Pattern Recognition (CVPR)*, pages 1637–1645, 2016. 4

[18] D. Kingma and J. Ba. Adam: A method for stochastic optimization. *International Conference on Learning Representations (ICLR)*, 2015. 4

[19] N. Kong, Y.-W. Tai, and J. S. Shin. A physically-based approach to reflection separation: from physical modeling to constrained optimization. *IEEE Transactions on Pattern Analysis and Machine Intelligence (T-PAMI)*, 36(2):209–221, 2014. 1, 2

[20] A. Krizhevsky, I. Sutskever, and G. E. Hinton. Imagenet classification with deep convolutional neural networks. In *Advances in Neural Information Processing Systems (NIPS)*, pages 1097–1105, 2012. 1

[21] A. Levin and Y. Weiss. User assisted separation of reflections from a single image using a sparsity prior. *IEEE Transactions on Pattern Analysis and Machine Intelligence (T-PAMI)*, 29(9), 2007. 1, 2, 3

[22] Y. Li and M. S. Brown. Exploiting reflection change for automatic reflection removal. In *IEEE International Conference on Computer Vision (ICCV)*, pages 2432–2439, 2013. 1, 2, 3, 6

[23] Y. Li and M. S. Brown. Single image layer separation using relative smoothness. In *IEEE Conference on Computer Vision and Pattern Recognition (CVPR)*, pages 2752–2759, 2014. 2, 4, 5, 6, 7

[24] S. Liu, J. Pan, and M.-H. Yang. Learning recursive filters for low-level vision via a hybrid neural network. In *European Conference on Computer Vision (ECCV)*, 2016. 1, 2, 7, 8

[25] T. Narihira, M. Maire, and S. X. Yu. Direct intrinsics: Learning albedo-shading decomposition by convolutional regression. In *IEEE International Conference on Computer Vision (CVPR)*, pages 2992–2992, 2015. 4

[26] S. Paris and F. Durand. A fast approximation of the bilateral filter using a signal processing approach. In *European Conference on Computer Vision (ECCV)*, pages 568–580, 2006. 1, 6

[27] J. S. Ren and L. Xu. On vectorization of deep convolutional neural networks for vision tasks. *AAAI Conference on Artificial Intelligence*, 2015. 1

[28] O. Russakovsky, J. Deng, H. Su, J. Krause, S. Satheesh, S. Ma, Z. Huang, A. Karpathy, A. Khosla, M. Bernstein, A. C. Berg, and L. Fei-Fei. ImageNet Large Scale Visual Recognition Challenge. *International Journal of Computer Vision (IJCV)*, 115(3):211–252, 2015. 1

[29] B. Sarel and M. Irani. Separating transparent layers through layer information exchange. In *European Conference on Computer Vision (ECCV)*, pages 328–341, 2004. 2, 4

[30] Y. Y. Schechner, N. Kiryati, and R. Basri. Separation of transparent layers using focus. *International Journal of Computer Vision (IJCV)*, 39(1):25–39, 2000. 1, 2

[31] Y. Y. Schechner, J. Shamir, and N. Kiryati. Polarization and statistical analysis of scenes containing a semireflector. *JOSA A*, 17(2):276–284, 2000. 1, 2

[32] Y. Shih, D. Krishnan, F. Durand, and W. T. Freeman. Reflection removal using ghosting cues. In *IEEE Conference on Computer Vision and Pattern Recognition (CVPR)*, pages 3193–3201, 2015. 2

[33] S. N. Sinha, J. Kopf, M. Goesele, D. Scharstein, and R. Szeliski. Image-based rendering for scenes with reflections. *ACM Transactions on Graphics (TOG)*, 31(4):100–1, 2012. 2

[34] R. Szeliski, S. Avidan, and P. Anandan. Layer extraction from multiple images containing reflections and transparency. In *IEEE Conference on Computer Vision and Pattern Recognition (CVPR)*, volume 1, pages 246–253, 2000. 2, 4

[35] R. Wan, B. Shi, T. A. Hwee, and A. C. Kot. Depth of field guided reflection removal. In *Image Processing (ICIP), 2016 IEEE International Conference on*, pages 21–25. IEEE, 2016. 2

[36] S. Xie and Z. Tu. Holistically-nested edge detection. In *International Conference on Computer Vision (ICCV)*, pages 1395–1403, 2015. 3

[37] L. Xu, C. Lu, Y. Xu, and J. Jia. Image smoothing via $L_0$ gradient minimization. In *ACM Transactions on Graphics (TOG)*, volume 30, page 174, 2011. 1, 5, 6, 8

[38] L. Xu, J. S. Ren, Q. Yan, R. Liao, and J. Jia. Deep edge-aware filters. In *International Conference on Machine Learning (ICML)*, pages 1669–1678, 2015. 1, 2, 3, 7, 8

[39] L. Xu, Q. Yan, Y. Xia, and J. Jia. Structure extraction from texture via natural variation measure. *ACM Transactions on Graphics (TOG)*, 2012. 1, 6

[40] T. Xue, M. Rubinstein, C. Liu, and W. T. Freeman. A computational approach for obstruction-free photography. *ACM Transactions on Graphics (TOG)*, 34(4):79, 2015. 1, 2, 3

[41] J. Yang, H. Li, Y. Dai, and R. T. Tan. Robust optical flow estimation of double-layer images under transparency or reflection. In *IEEE Conference on Computer Vision and Pattern Recognition (CVPR)*, pages 1410–1419, 2016. 1, 2, 4

[42] Q. Zhang, X. Shen, L. Xu, and J. Jia. Rolling guidance filter. In *European Conference on Computer Vision (ECCV)*, pages 815–830, 2014. 1, 6

[43] Q. Zhang, L. Xu, and J. Jia. 100+ times faster weighted median filter. In *IEEE Conference on Computer Vision and Pattern Recognition (CVPR)*, 2014. 1, 8

# A Generic Deep Architecture for Single Image Reflection Removal and Image Smoothing
## (*Supplementary Material*)

## 1 Outline

This supplementary document provides more analyses and results that were not included in the main paper due to space limitations. The contents are organized as follows.

- **Section 2:** Detailed explanation and analysis of our synthetic data generation method for training CEILNet in the reflection removal task.
- **Section 3:** Comparison of an I-CNN network with varying convolutional layers against the full CEILNet pipeline.
- **Section 4:** Visualization of more edge maps predicted by E-CNN for both the reflection removal and image smoothing tasks.
- **Section 5:** More visual comparisons with existing deep image smoothing networks.
- **Section 6:** More visual comparisons with our baselines and existing methods on the reflection removal problem.
- **Section 7:** More visual results of our CEILNet on both synthetic and real reflection images. For the latter, our network is the only approach capable of reliably separating a series of difficult single real-world images into reflection and background layers without strong additional priors.
- **Section 8:** More visual results of our CEILNet for different smoothing filters.

# 2 Complete description of our synthetic reflection image generation process

Real images with ground truth background and reflection layers are difficult to obtain. In the main paper, we have proposed a novel reflection image data generation method for producing synthetic training data. Here we first elaborate on why simply mixing two images does not work well for this purpose, followed by a thorough explanation of how our more involved generation strategy operates.

## 2.1 The problem with naively mixing two images

To generate enough data for training a deep network, a naive approach is to simply mix a candidate background layer $\mathbf{B}$ and reflection layer $\mathbf{R}$ to create new synthetic image samples via $\mathbf{I} = \nu_1\mathbf{B} + \nu_2\mathbf{R}$, where $\nu_1$ and $\nu_2$ are relative scaling coefficients that sum to one to avoid overflow or image clipping (*e.g.*, 0.7 and 0.3 respectively). Indeed this exact strategy has been widely used in previous works to create ground-truth for analysis and quantitative evaluation [9, 7, 3, 5, 14]. But as a viable source of training data, this approach is problematic based on several inconsistencies with natural images:

- First, both $\mathbf{B}$ and $\mathbf{R}$ should not be scaled. In real-world images, background and reflection layers contain various levels of luminance, from the darkest to the brightest color. Scaling the images not only constrains each layer within a relatively smaller color range, but also suppresses abrupt color transitions, especially for the reflection layer.
- Second, in typical images the reflection layer will only partially cover the background. In fact, the visibility of the reflection layer depends on the relative intensity between the transmitted light from the background scene and the reflected light. Hence we often observe large regions where no reflection is visible at all, even for scenes viewed entirely through a window or other glass surface.

In Section 6 below we demonstrate that training with such linearly mixed images does not perform well.

## 2.2 Details of our alternative generation method

Our proposed pipeline is annotated in Figure 1, with an illustrative example at the top and detailed explanations of each step shown below. The crux of our method is a heuristic and simple subtraction operation: to simulate real-life reflection images, we sum up an unmodified natural image $\mathbf{B}$ and another attenuated natural image $\mathbf{R}$, which is subtracted by one single, adaptively-computed value across the whole image (see Step 4 in Figure 1).

The advantages of this subtraction strategy (which replaces any scaling operation) can be summarized as follows:

- First, the oversaturation side effects in $\mathbf{I}$ generated by directly adding up $\mathbf{B}$ and $\tilde{\mathbf{R}}$ are eliminated mostly; see the difference of generated reflection images $\mathbf{I}$ in Step 2 and Step 6 of Figure 1.
- Second, large gradients or abrupt color transitions in the original reflection layer are well maintained, as subtracting a scalar from the whole image does not affect luminance differences while the scaling operation used by naive mixing does.
- Third, strong reflections can occur as in real-world cases, especially when the given background layer is relatively dark such that oversaturation is insignificant and the subtracted part of the reflection in Step 4 (which is proportional to the mean) is small. This adaptively enables a wider color range for the reflection layer when the background is weak.
- Lastly, reflection-free regions can also naturally arise, since the subtraction and color clipping may lead to zero brightness in the reflection layer.

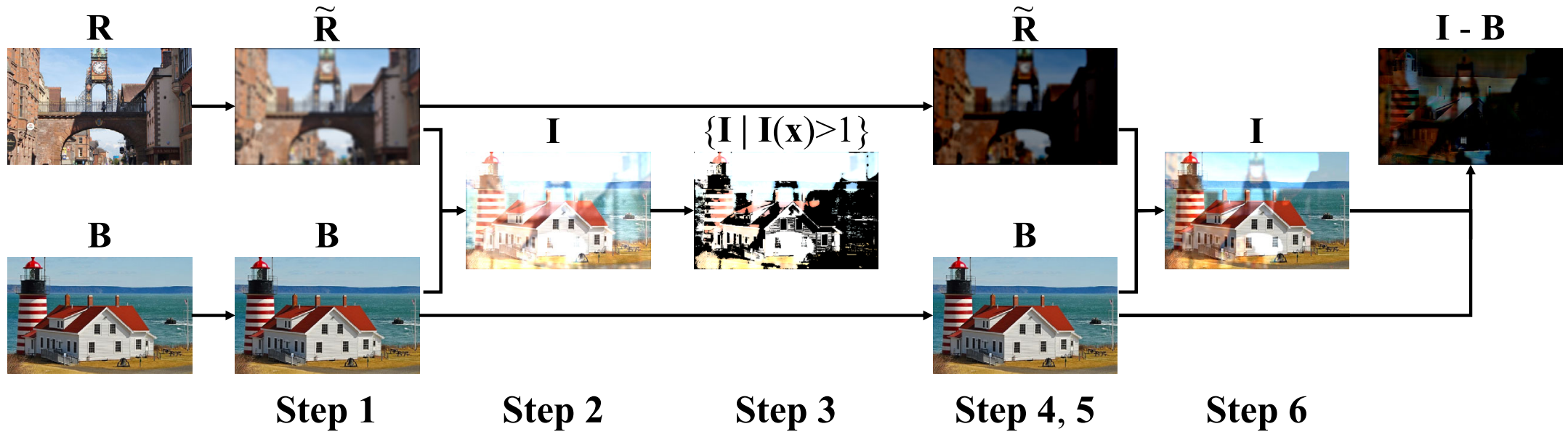

Randomly pick two natural images normalized to $[0, 1]$ as background layer $\mathbf{B}$ and reflection layer $\mathbf{R}$ respectively, then:

**Step 1.** $\tilde{\mathbf{R}} \leftarrow gauss_blur_{\sigma}(\mathbf{R})$ with $\sigma \sim \mathcal{U}(2, 5)$

$\triangleright$ *Blur the reflection layer with a Gaussian filter. The standard deviation of the Gaussian filter is randomly picked in a wide range of* $[2, 5]$ *to accommodate more diversity in the reflection layer.*

**Step 2.** $\mathbf{I} \leftarrow \mathbf{B} + \tilde{\mathbf{R}}$

$\triangleright$ *Tentatively generate the synthesized image by directly adding up the background layer and the (blurred) reflection layer.*

**Step 3.** $m \leftarrow mean(\{\mathbf{I}(\mathbf{x}, c) \mid \mathbf{I}(\mathbf{x}, c) > 1, \forall \mathbf{x}, \forall c = 1,2,3\})$

$\triangleright$ *Calculate the mean of all the color (R, G, B) values in the synthesized image that are out of color range (i.e., above 1).*

**Step 4.** $\tilde{\mathbf{R}}(\mathbf{x}, c) \leftarrow \tilde{\mathbf{R}}(\mathbf{x}, c) - \gamma \cdot (m-1), \forall \mathbf{x}, \forall c$

$\triangleright$ *Reduce the color (R, G, B) values of the reflection layer by the calculated mean (the overflowed fraction only; hence* $m - 1$*). We also magnify the subtracted part by a scalar* $\gamma$ *set as 1.3 to further attenuate the reflection. We found that this heuristic for reflection layer attenuation works well in practice (and our networks are trained exclusively via data obtained this way), although there may exist other possible choices.*

**Step 5.** $\tilde{\mathbf{R}} \leftarrow clip_{[0,1]}(\tilde{\mathbf{R}})$

$\triangleright$ *Clip the color values of the reflection layer to the valid range (to deal with negative values arisen due to the subtraction in Step. 4).*

**Step 6.** $\mathbf{I} \leftarrow clip_{[0,1]}\big(\mathbf{B} + \tilde{\mathbf{R}}\big)$

$\triangleright$ *Generate the final image by adding up the background layer and the modified reflection layer followed by clipping the color to the valid range (to deal with values above 1).*

Output $\mathbf{I}$ as the synthesized image with $\mathbf{B}$ as the ground-truth background layer, and $\mathbf{I} - \mathbf{B}$ as the ground-truth reflection layer.

Figure 1: Our reflection image data synthesis pipeline for weakly-supervised learning. We illustrate the process with an example (*top*), followed by detailed step-by-step explanations (*bottom*).

# 3 Performance using an I-CNN without predicted edges

In Section 5.1 of the main paper, we described the performance of a single I-CNN (which is equivalent to our CEILNet pipeline without the E-CNN and edge supervision) as more convolution layers are stacked. Figure 2 presents the detailed results. As can be seen, using I-CNN only the performance gets saturated very quickly, and the best performing 70 layer I-CNN (PSNR 33.37 dB) lags far behind the CEILNet (PSNR 37.10 dB), even though the latter has fewer parameters (CEILNet has basically the same number of parameters as a 64 layer I-CNN).

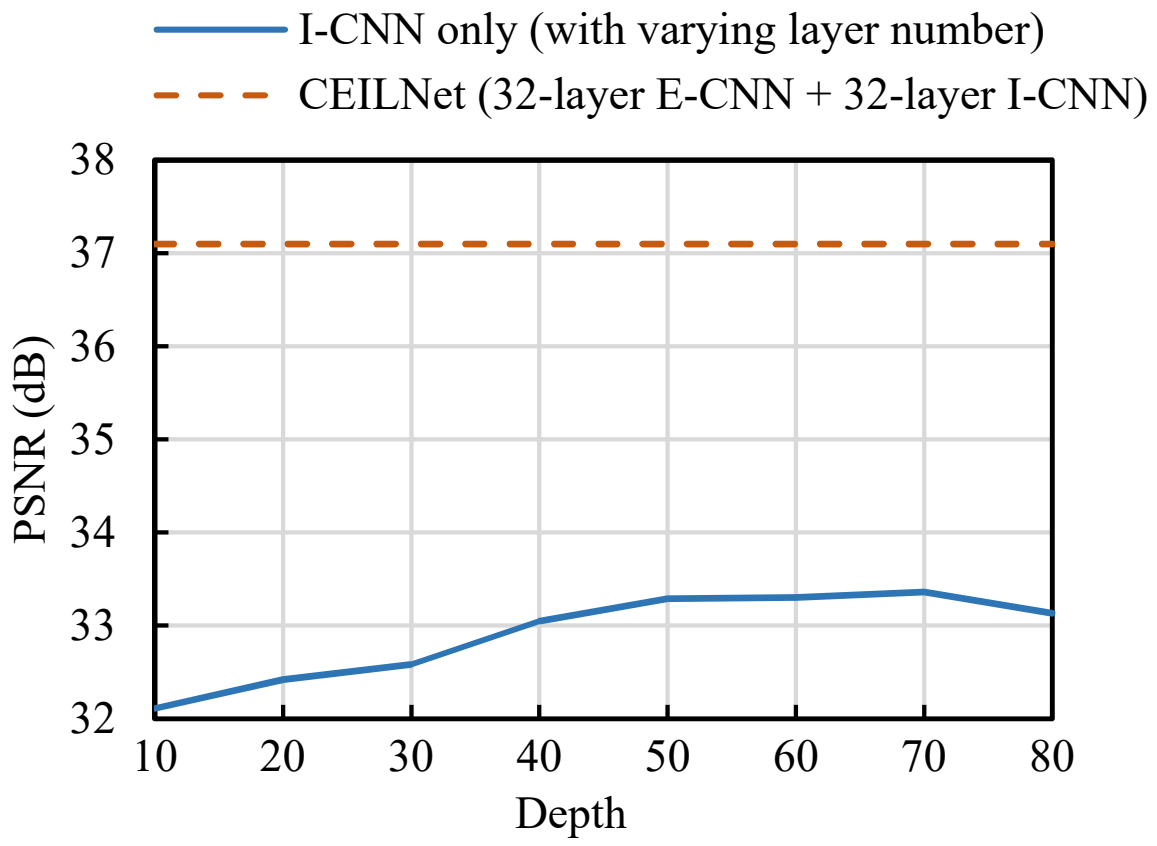

Figure 2: Performance of a simple I-CNN network with respect to network depth for the task of learning an $L_0$ smoothing filter [10]. When the depth is 64, the I-CNN has basically the same number of parameters as our proposed CEILNet. Clearly, the benefit of the latter is not just a larger parameter set, but rather its purposeful integration with the E-CNN.

# 4 Edge map visualization and analysis

In this section, we present more edge maps generated by our E-CNN in both image smoothing (Figure 3 top two rows) and reflection removal (Figure 3 bottom two rows).

In the image smoothing task, fundamental image constituents, *e.g.*, salient edges, will be maintained, and insignificant details will be diminished. In our method, the edge map of the target image is predicted by E-CNN, then used to process the input image and generate the result with I-CNN. The two examples in Figure 3 show that our E-CNN is able to remove the insignificant details, meanwhile preserving prominent edges and rendering them visually more distinct. With the guidance of the predicted edge map, I-CNN generates high fidelity smoothing results compared to the original filter (RTV [12] in these examples).

For separating reflection and background layers, edges can still play an important role to differentiate the two independent layers [4, 13, 14]. Different from image smoothing, in this task the desired edge map is from the entire image structure of the background layer only. As shown in the bottom two rows of Figure 3, our E-CNN can remove the edges belonging to reflections and I-CNN can reconstruct clean background images. Comparison of the results with and without edge prediction can be found in Section 6 below.

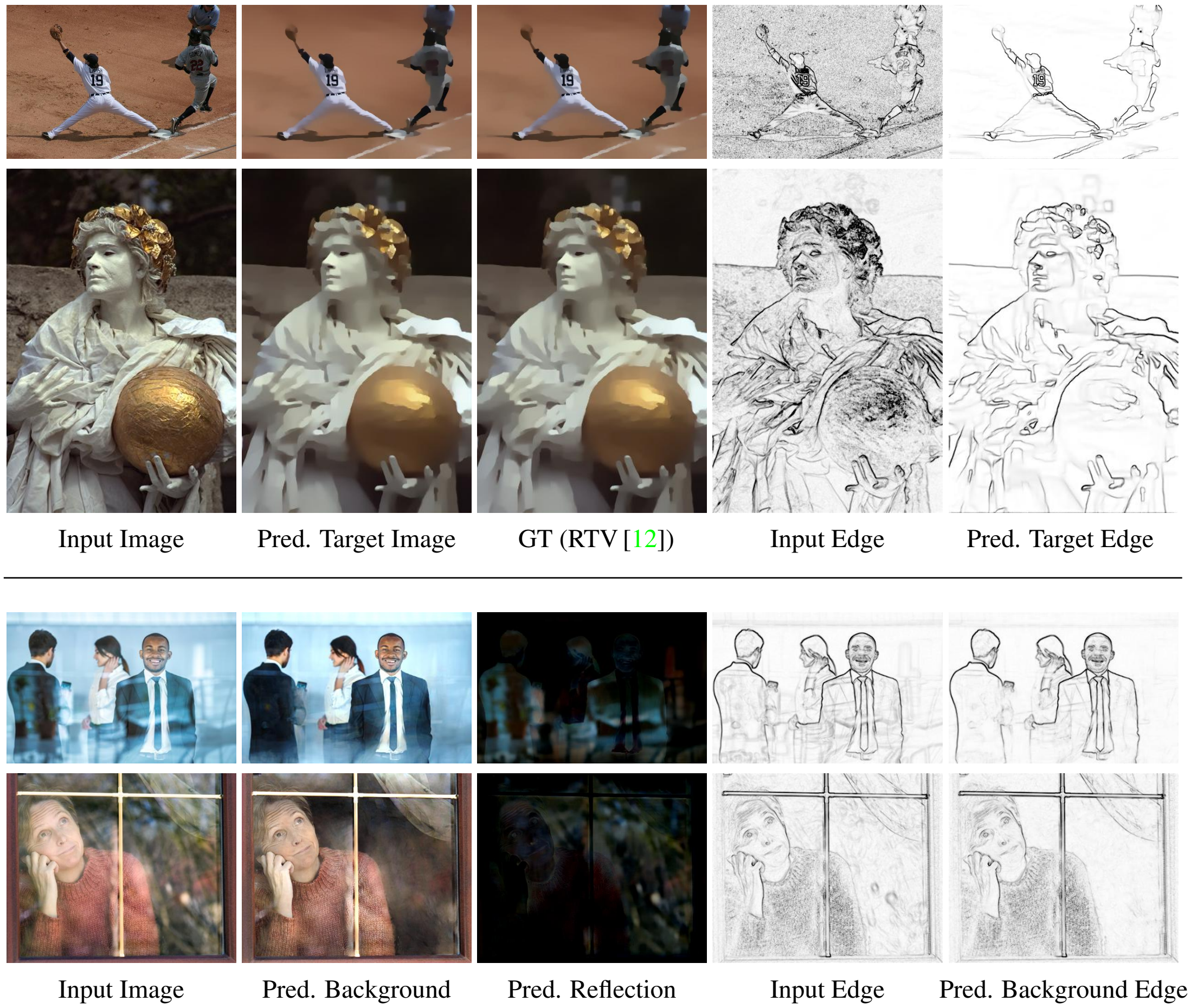

Figure 3: Visualization of the predicted edge maps in the image smoothing (top two rows) and reflection removal (bottom two rows) tasks. GT denotes ground truth.

# 5 Detailed comparison of different deep image smoothing networks

In the main paper, we have compared with existing deep image smoothing methods [11, 6] quantitatively and qualitatively. Here we analyze these methods and compare them with ours in more details. Additional qualitative comparisons are also provided in Figure 4.

The approach from [11] predicts the gradient map of the target image using a shallow network (3 convolution layers), and reconstructs the smoothed images by solving a separate optimization problem. One potential drawback of this method is that the optimization step completely relies on the predicted gradient map, and the gradient errors can easily spread to a surrounding area (see the boundary of "bull" and "building" cases in Figure 4). Our predicted edge map also represents sparse salient structures similar to the gradient map in [11]. But unlike [11], our image reconstruction is learned by the deep neural network, which exhibits higher robustness to errors of the predicted edge map.

The method of [6] generates smoothed images using separable recursive 1D filters based on the predicted weight maps by a CNN. Their weight map serves a similar role with a gradient map; however, it seems impracticable that one single (parameter-free) filtering technique in [6] can well approximate various existing edge-aware filters of different effects and disparate algorithm details, even if the weight map in [6] is generated by a CNN trained for each filter. As can be seen in the row of Figure 4, their learned $L_0$ filter turns out to have some obvious artifacts. In contrast, our filtering process is based on a network with parameters trained for each smoothing algorithm, which can generate more accurate approximations for different smoothing algorithms.

# 6 Detailed comparison with our baselines and previous methods on the reflection removal task

In the main paper (Figure 5), we have presented one real image result comparison between our CEILNet and two of its baselines: 1) CEILNet trained using naïvely generated reflection images and 2) I-CNN only (*i.e.*, without the predicted edge from E-CNN). We now provide more explanations and comparisons as follows. By "naïve", we mean linearly combining the background and reflection layer with two constant coefficients that sum up to 1 (see also Section 2.1). In our experiments, the coefficients vary in a wide range to account for different situations: $[0.6, 0.9]$ for $\mathbf{B}$ (thus $[0.1, 0.4]$ for $\mathbf{R}$). As shown in Figure 5, CEILNet-naïve can hardly remove real-life reflections; its results are clearly not comparable to CEILNet, which demonstrated the effectiveness of our proposed training image synthesis method. The results from an I-CNN only are better than CEILNet-naïve with more reflections removed, but are still clearly inferior compared to CEILNet. The advantage of our edge prediction is therefore also demonstrated.

The method of [5], perhaps the most closely related algorithm to ours[1], shares some similar properties as CEILNet-naïve. It assumes that (*i*) it is less likely to have abrupt color transition in the reflection layer, and (*ii*) reflection is almost everywhere in the image (*i.e.*, reflection-free regions are rare). As can been seen in the last row of Figure 5, it tends to extract reflections that are very blurry and cover the whole image. Note that the fourth example in (d) is an image collected from [5]. On such cases where the assumptions from [5] are valid, our method performs comparably with [5]. But on other cases that are more difficult (and yet more common, such as other images in Figure 5 and the images in Section 7), our method excels.

[1]The method of [8] also performs reflection removal from a single image automatically like ours. However, their method is restricted to images that contain ghost effects with two ghost layers. For reflection images beyond this limited scope (such as most of the images in our main paper and this supplementary file), their method does not work at all.

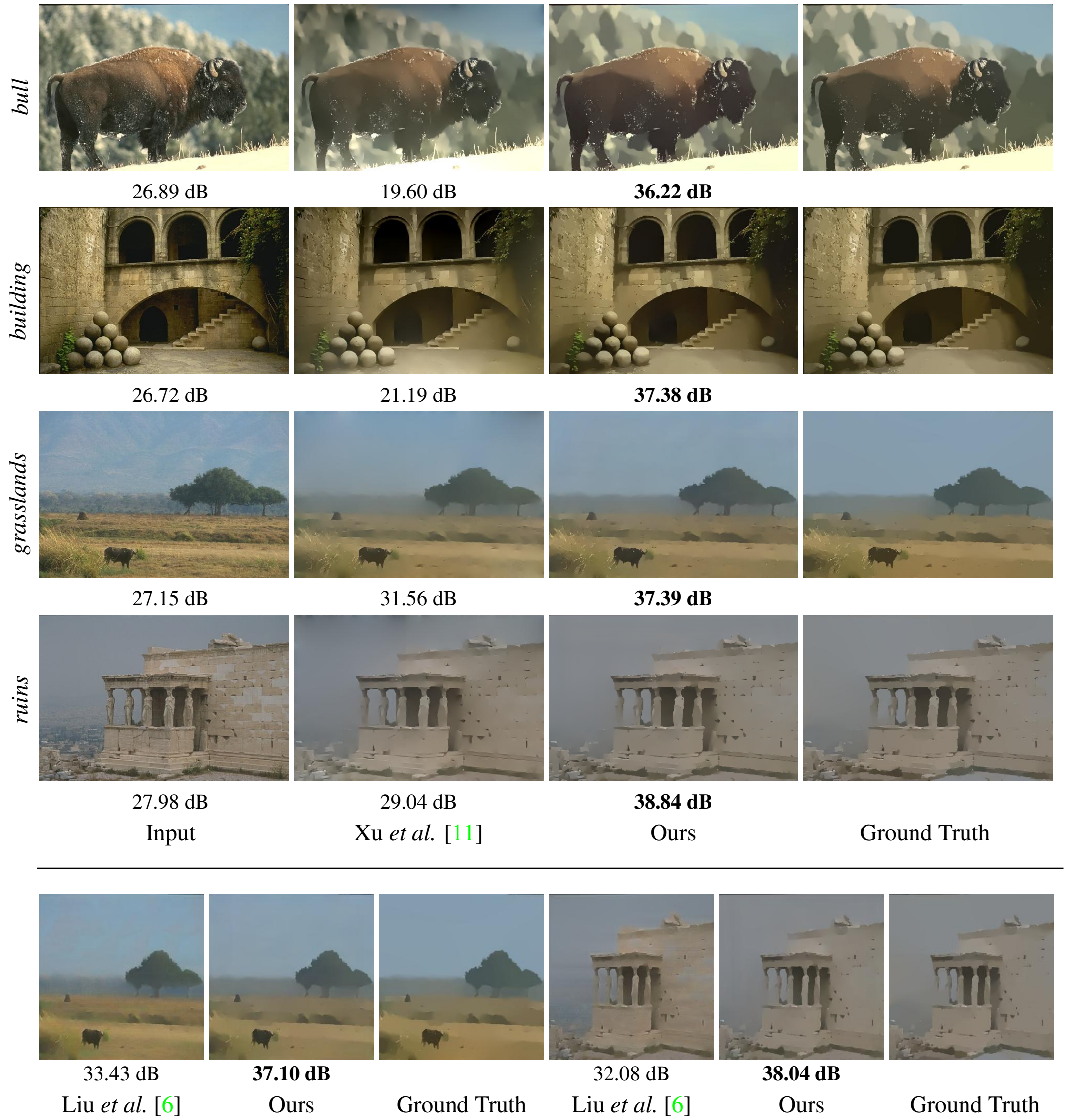

Figure 4: Qualitative comparison on the image smoothing task. All the methods are trained to approximate $L_0$ smoothing [10]. Top: Comparison with Xu *et al.* [11]. Bottom: Comparison with Liu *et al.* [6] on the 256×256 image size. Our results are visually much closer to the ground truth. The numbers show the PSNR values.

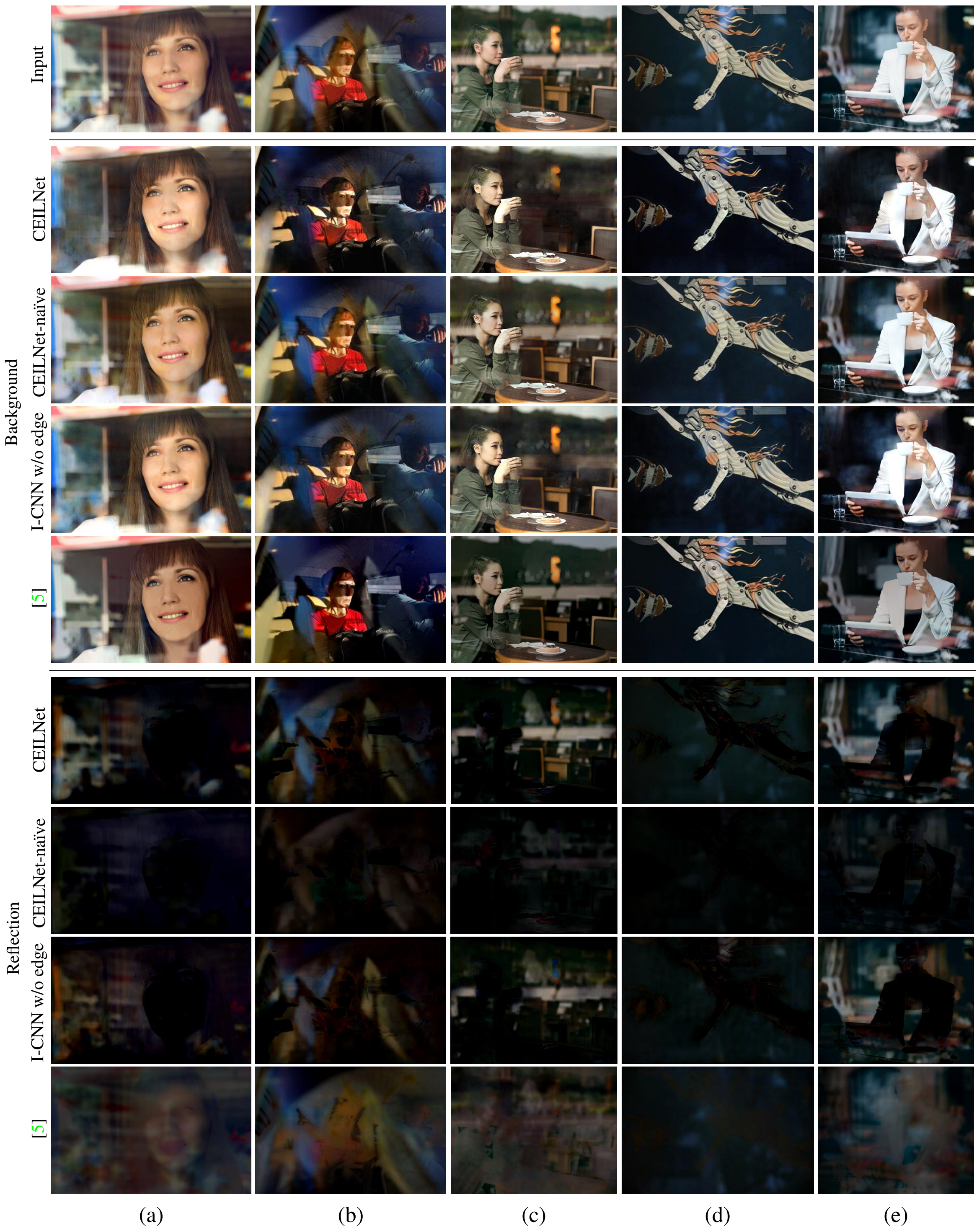

Figure 5: Qualitative comparison on real reflection images with our baselines and Li and Brown [5].

## 7 More results for reflection removal

This section presents more qualitative results on synthetic reflection images (Figure 6, 7) and real-world photographs (Figure 8, 9, 10, 11).

We emphasize that, because no existing database of real refection images exists with ground truth, in all cases our model was trained solely using images generated synthetically via the process described in Section 2. Moreover, given the extreme difficulty of separating a single image into background and reflection layers, no existing algorithm of any kind succeeds on these real-world cases. Nonetheless, our approach still produces reasonable results in most situations, *i.e.*, even when reflections cannot be completely removed, the recovered background images are significantly clean relative to the original.

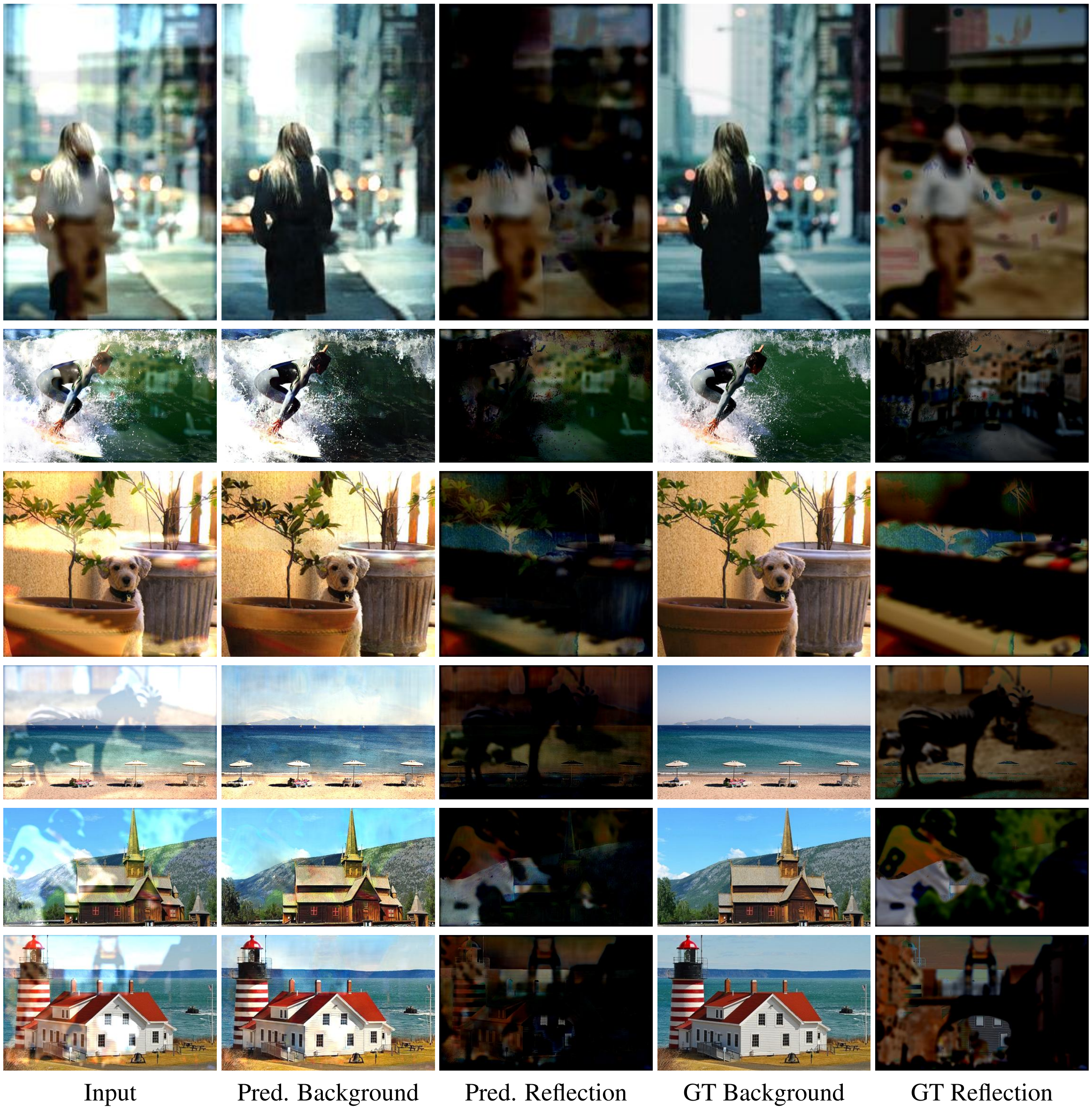

Figure 6: More visual results of our CEILNet on synthetic reflection images. GT denotes ground truth.

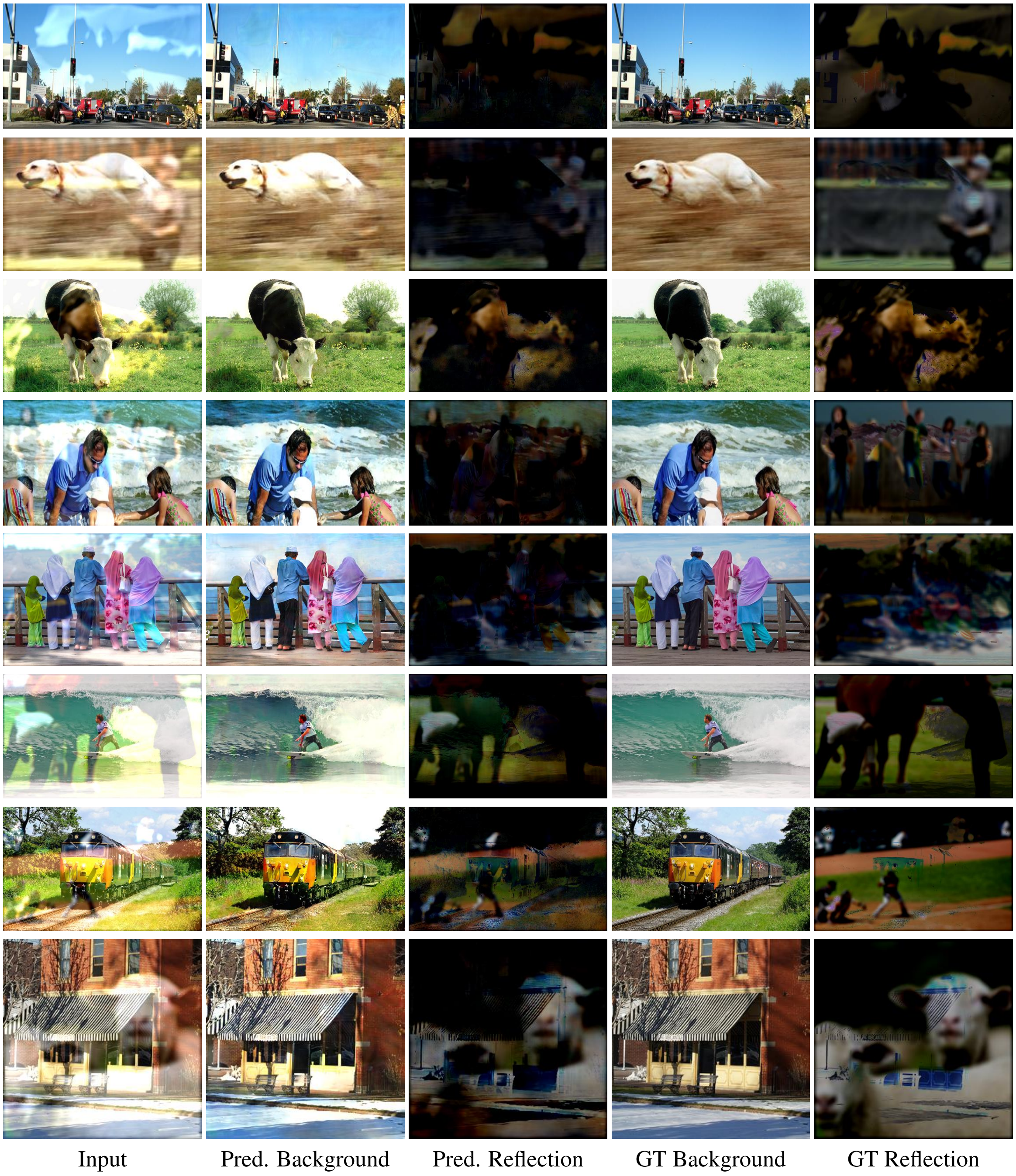

Figure 7: More visual results of our CEILNet on synthetic reflection images. GT denotes ground truth.

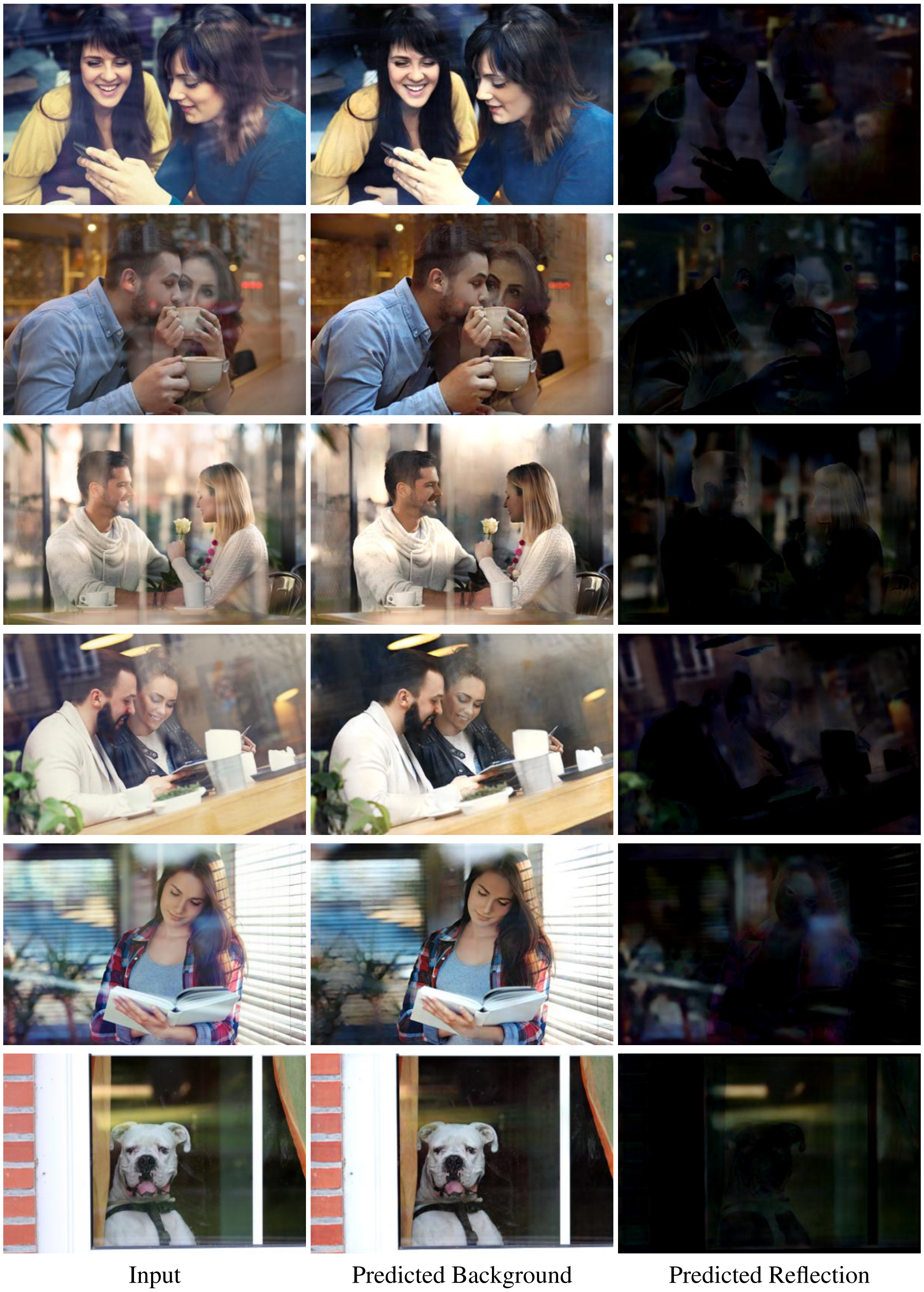

Figure 8: More visual results of our CEILNet on real reflection images.

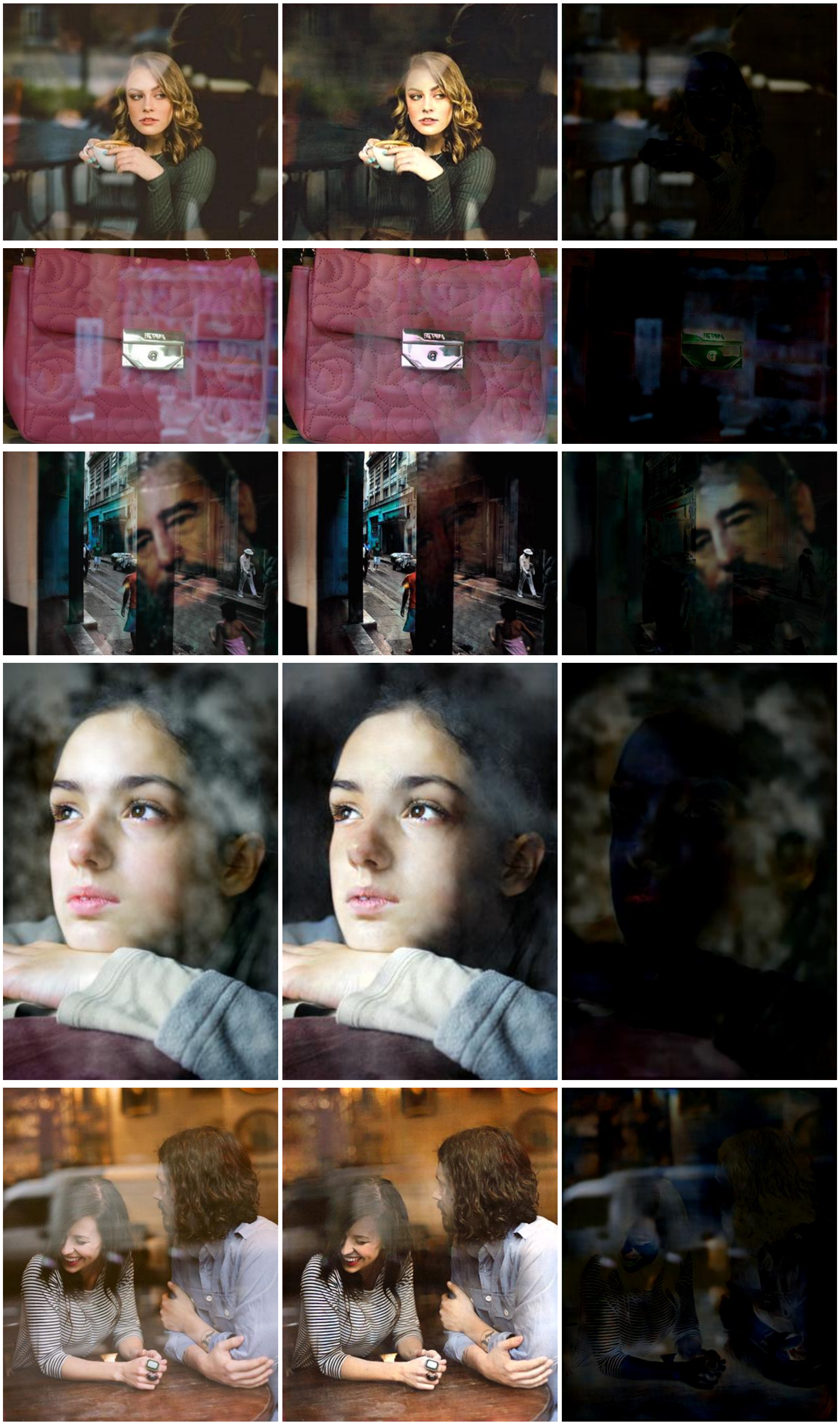

Input Predicted Background Predicted Reflection

Figure 9: More visual results of our CEILNet on real reflection images.

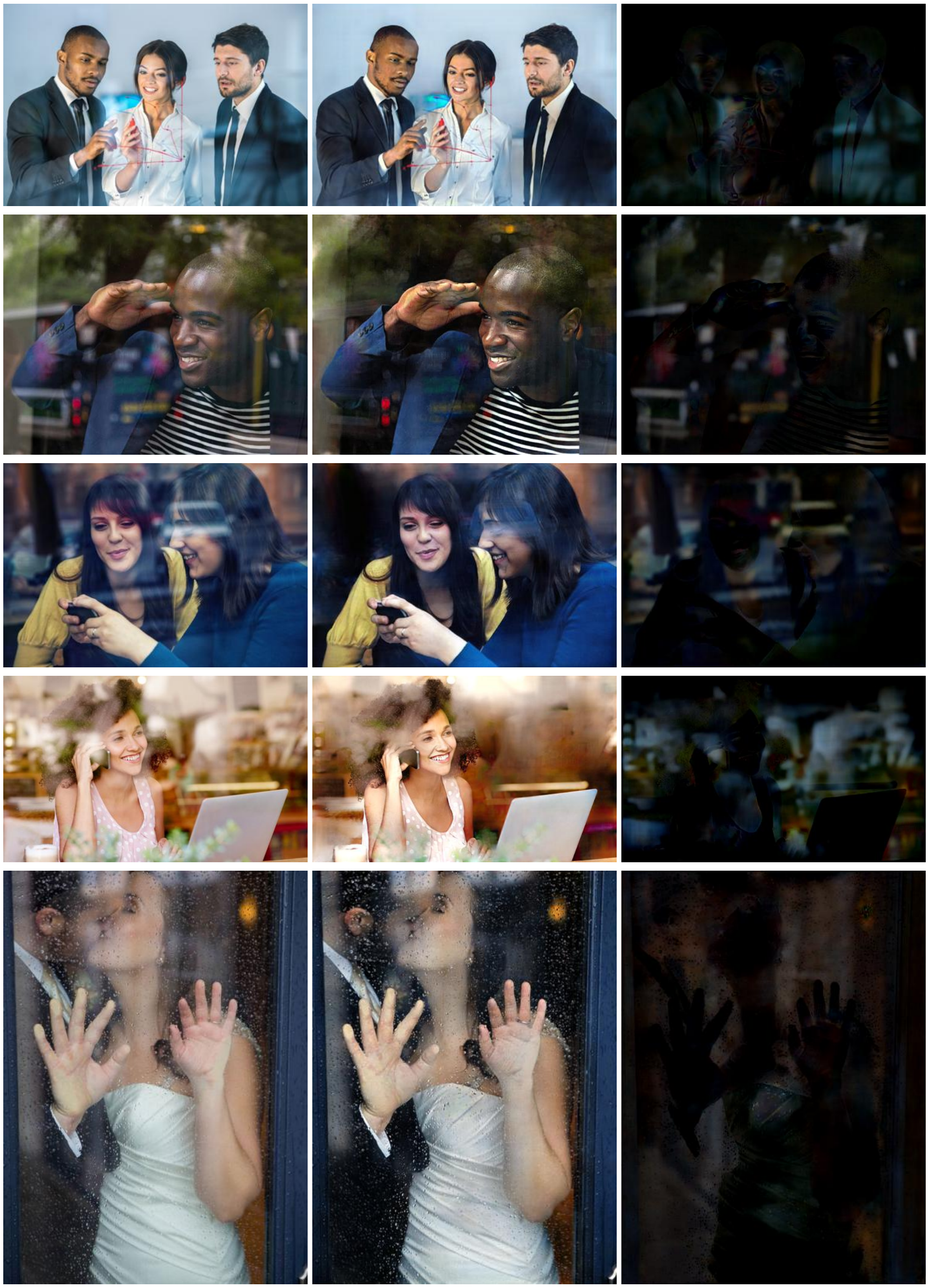

Input Predicted Background Predicted Reflection

Figure 10: More results of our CEILNet on real reflection images.

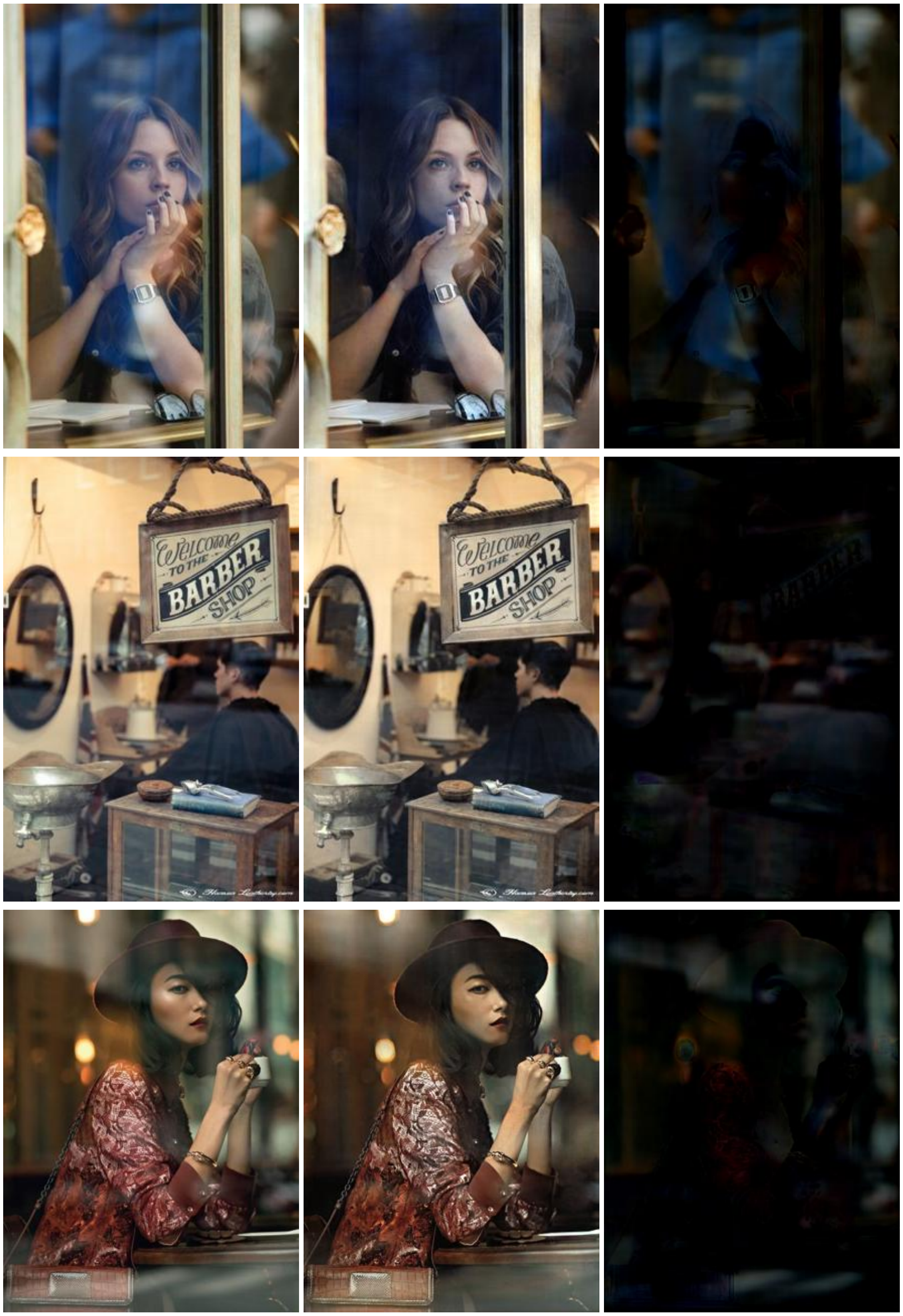

Input | Predicted Background | Predicted Reflection

Figure 11: More visual results of our CEILNet on real reflection images.

# 8 More results for image smoothing

In this section, we present more visual results for the image smoothing tasks (Figure 12, 13, 14, 15 16 for approximating $L_0$ [10], $L_1$ [1], RTV [12], RGF [15], WLS [2] filters respectively).

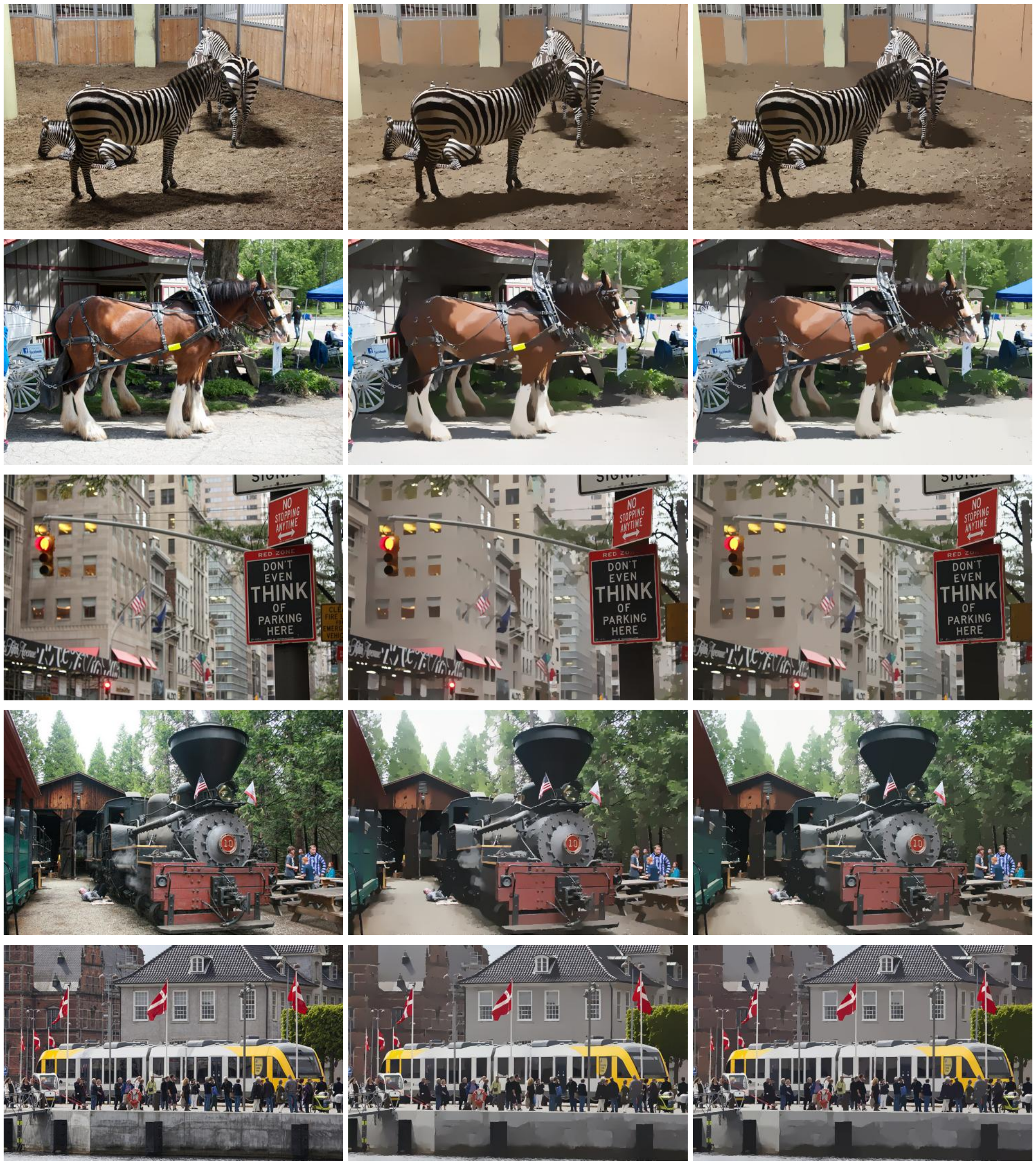

Input | Our Result | Ground Truth ($L_0$ [10])

Figure 12: Approximation of the $L_0$ image smoothing algorithm [10] using our CEILNet.

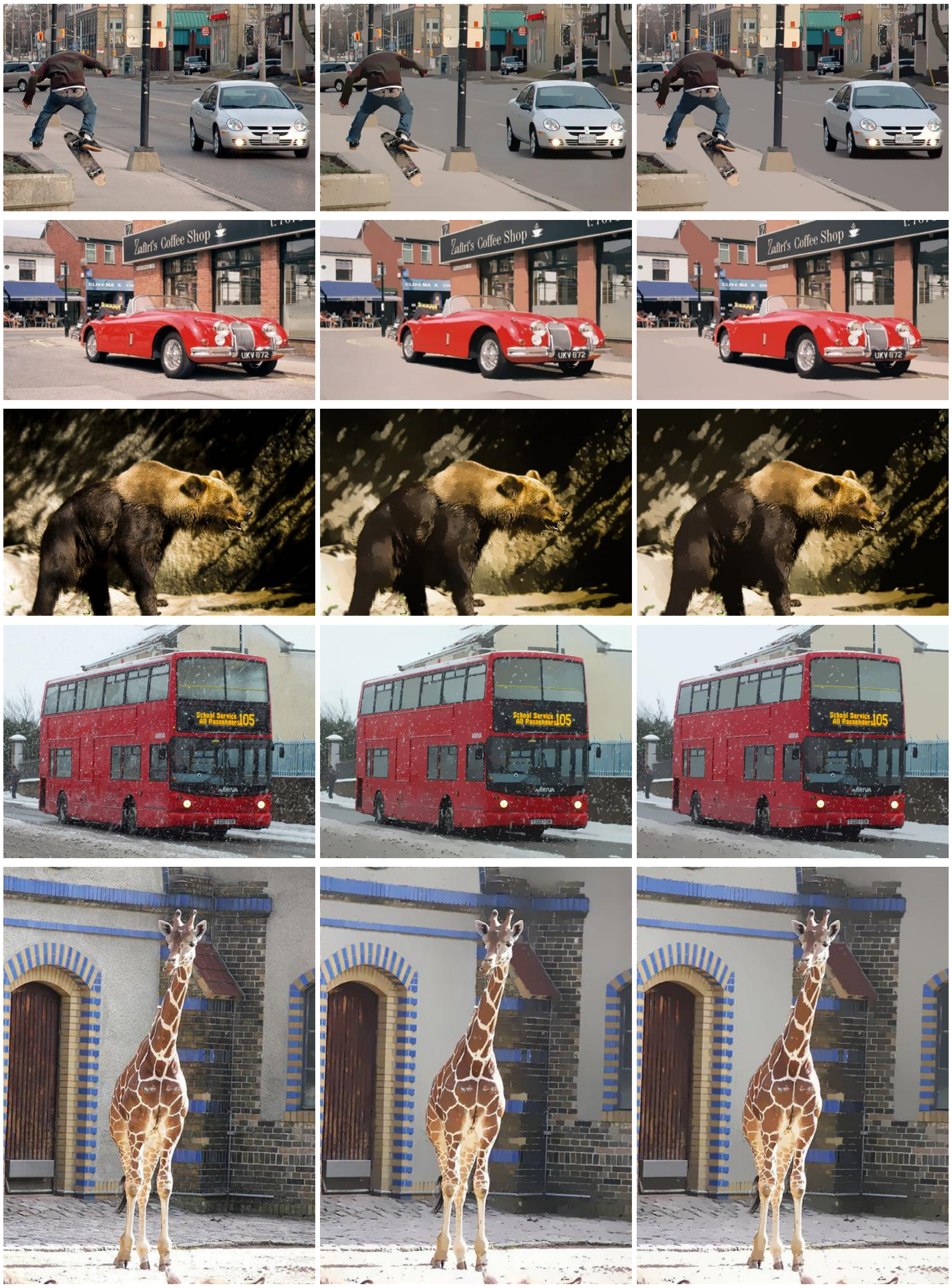

Input | Our Result | Ground Truth ($L_1$ [1])

Figure 13: Approximation of the $L_1$ image smoothing algorithm [1] using our CEILNet.

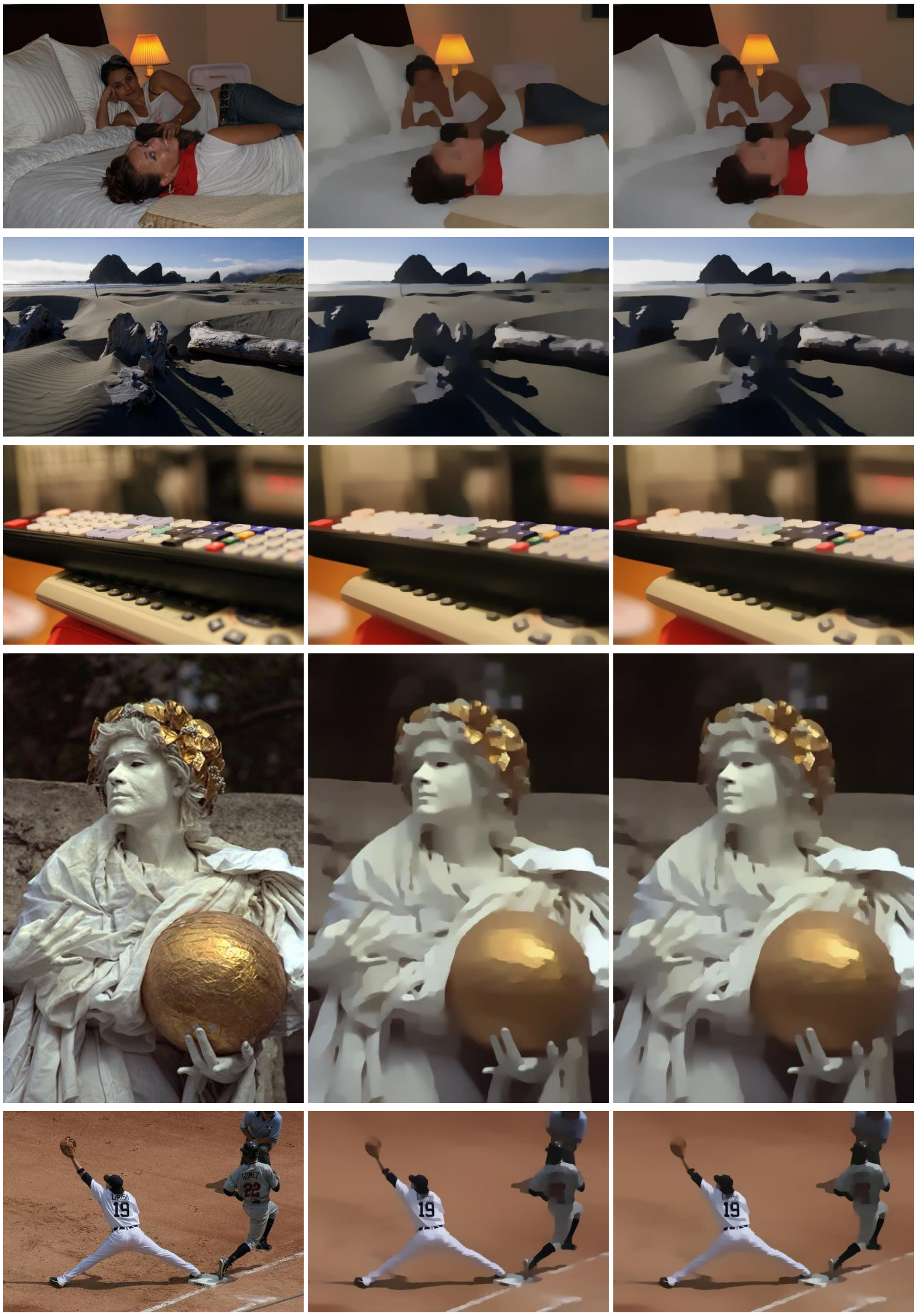

Input | Our Result | Ground Truth (RTV [12])

Figure 14: Approximation of the RTV image smoothing algorithm [12] using our CEILNet.

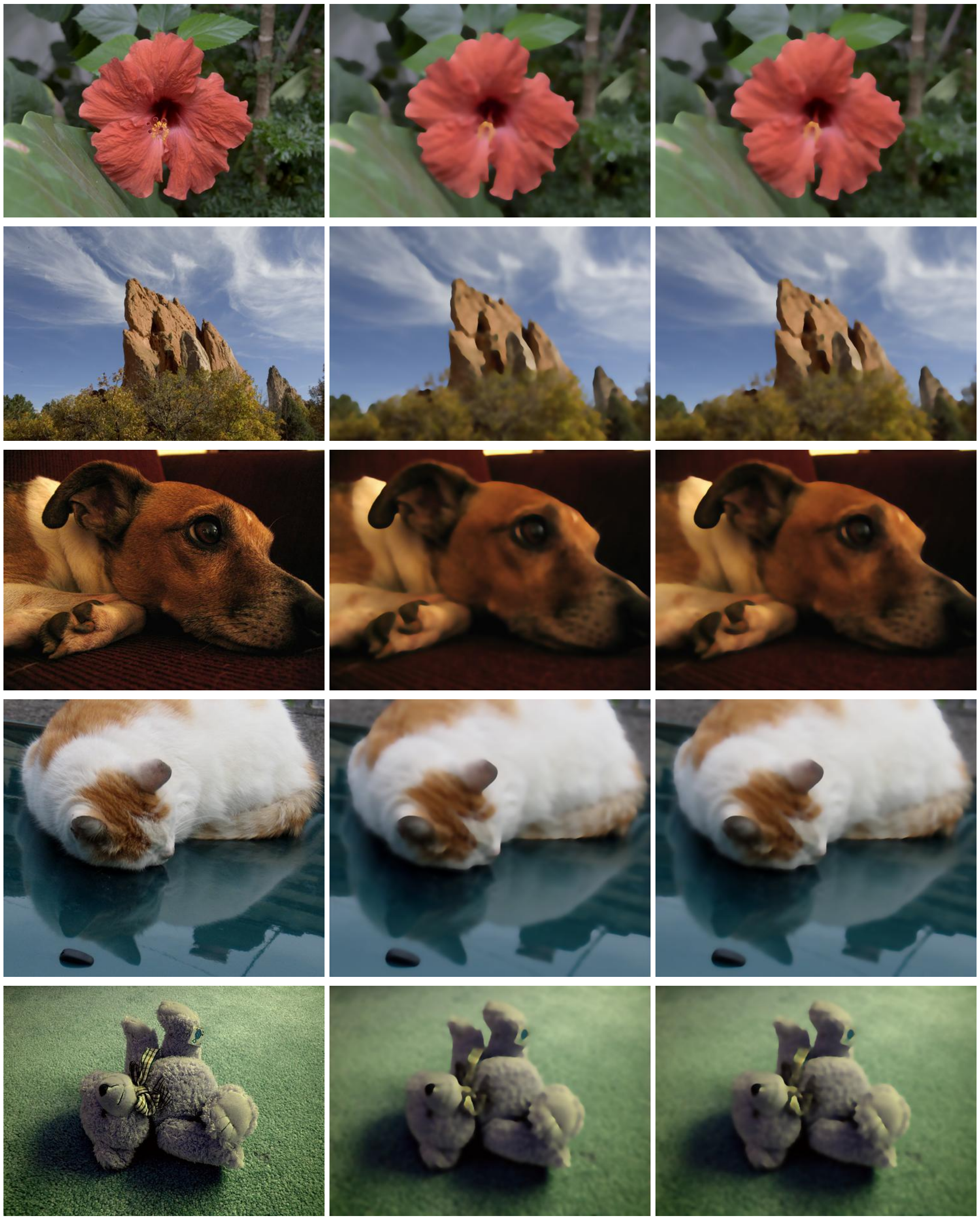

Figure 15: Approximation of the RGF image smoothing algorithm [15] using our CEILNet.

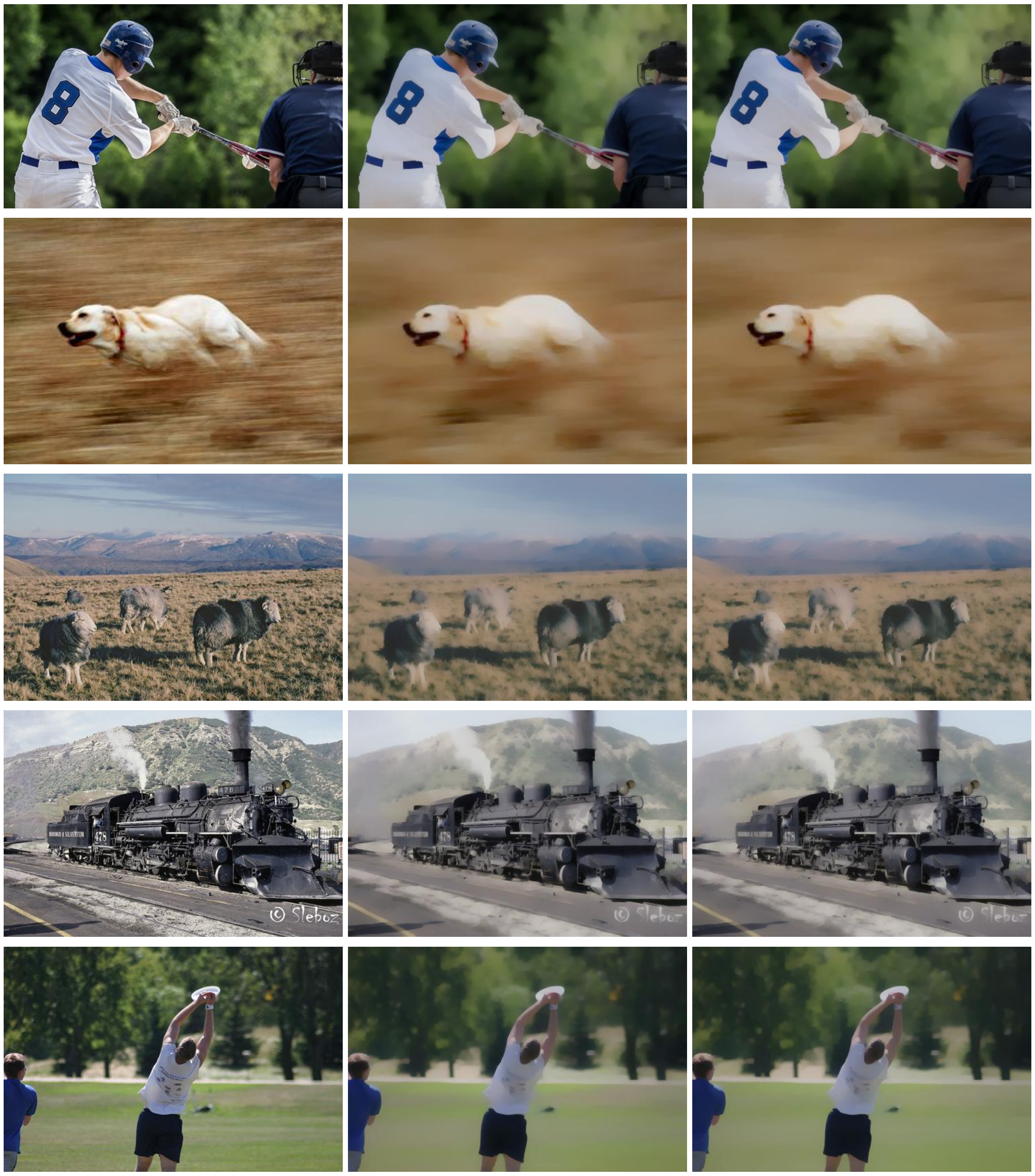

Input | Our Result | Ground Truth (WLS [2])

Figure 16: Approximation of the WLS image smoothing algorithm [2] using our CEILNet.

# References

[1] S. Bi, X. Han, and Y. Yu. An $L_1$ image transform for edge-preserving smoothing and scene-level intrinsic decomposition. *ACM Transactions on Graphics (TOG)*, 34(4):78, 2015. 15, 16

[2] Z. Farbman, R. Fattal, D. Lischinski, and R. Szeliski. Edge-preserving decompositions for multi-scale tone and detail manipulation. *ACM Transactions on Graphics (TOG)*, 27(3), 2008. 15, 19

[3] X. Guo, X. Cao, and Y. Ma. Robust separation of reflection from multiple images. In *IEEE Conference on Computer Vision and Pattern Recognition (CVPR)*, pages 2187–2194, 2014. 2

[4] Y. Li and M. S. Brown. Exploiting reflection change for automatic reflection removal. In *IEEE International Conference on Computer Vision (ICCV)*, pages 2432–2439, 2013. 5

[5] Y. Li and M. S. Brown. Single image layer separation using relative smoothness. In *IEEE Conference on Computer Vision and Pattern Recognition (CVPR)*, pages 2752–2759, 2014. 2, 6, 8

[6] S. Liu, J. Pan, and M.-H. Yang. Learning recursive filters for low-level vision via a hybrid neural network. In *European Conference on Computer Vision (ECCV)*, 2016. 6, 7

[7] B. Sarel and M. Irani. Separating transparent layers through layer information exchange. In *European Conference on Computer Vision (ECCV)*, pages 328–341, 2004. 2

[8] Y. Shih, D. Krishnan, F. Durand, and W. T. Freeman. Reflection removal using ghosting cues. In *IEEE Conference on Computer Vision and Pattern Recognition (CVPR)*, pages 3193–3201, 2015. 6

[9] R. Szeliski, S. Avidan, and P. Anandan. Layer extraction from multiple images containing reflections and transparency. In *IEEE Conference on Computer Vision and Pattern Recognition (CVPR)*, volume 1, pages 246–253, 2000. 2

[10] L. Xu, C. Lu, Y. Xu, and J. Jia. Image smoothing via $L_0$ gradient minimization. In *ACM Transactions on Graphics (TOG)*, volume 30, page 174, 2011. 4, 7, 15

[11] L. Xu, J. S. Ren, Q. Yan, R. Liao, and J. Jia. Deep edge-aware filters. In *International Conference on Machine Learning (ICML)*, pages 1669–1678, 2015. 6, 7

[12] L. Xu, Q. Yan, Y. Xia, and J. Jia. Structure extraction from texture via natural variation measure. *ACM Transactions on Graphics (TOG)*, 2012. 5, 15, 17

[13] T. Xue, M. Rubinstein, C. Liu, and W. T. Freeman. A computational approach for obstruction-free photography. *ACM Transactions on Graphics (TOG)*, 34(4):79, 2015. 5

[14] J. Yang, H. Li, Y. Dai, and R. T. Tan. Robust optical flow estimation of double-layer images under transparency or reflection. In *IEEE Conference on Computer Vision and Pattern Recognition (CVPR)*, pages 1410–1419, 2016. 2, 5

[15] Q. Zhang, X. Shen, L. Xu, and J. Jia. Rolling guidance filter. In *European Conference on Computer Vision (ECCV)*, pages 815–830, 2014. 15, 18